\pdfoutput=1

\PassOptionsToPackage{table}{xcolor}
\documentclass[11pt]{article}

\usepackage{acl}

\usepackage{times}
\usepackage{latexsym}

\usepackage{amsmath,amsfonts,bm}

\def\eqref#1{equation~\ref{#1}}

\def\1{\bm{1}}

\def\rvr{{\mathbf{r}}}

\def\rvy{{\mathbf{y}}}
\def\rvz{{\mathbf{z}}}

\def\va{{\bm{a}}}
\def\vb{{\bm{b}}}

\def\vx{{\bm{x}}}

\def\mR{{\bm{R}}}

\def\mX{{\bm{X}}}
\def\mY{{\bm{Y}}}
\def\mZ{{\bm{Z}}}

\DeclareMathAlphabet{\mathsfit}{\encodingdefault}{\sfdefault}{m}{sl}
\SetMathAlphabet{\mathsfit}{bold}{\encodingdefault}{\sfdefault}{bx}{n}

\def\sA{{\mathbb{A}}}
\def\sB{{\mathbb{B}}}

\def\sX{{\mathbb{X}}}
\def\sY{{\mathbb{Y}}}

\usepackage{booktabs}
\usepackage{amsfonts}
\usepackage{graphicx}
\usepackage{subcaption}
\usepackage{float}
\usepackage{multirow}
\usepackage[inline]{enumitem}
\usepackage{array}
\usepackage{amsmath}
\newcommand\Tstrut{\rule{0pt}{2.6ex}}         
\newcommand\Bstrut{\rule[-0.9ex]{0pt}{0pt}}   
\usepackage{multirow}
\usepackage{adjustbox}
\usepackage{enumitem}
\usepackage[section]{placeins}
\usepackage{hyperref}

\usepackage[tight-spacing=true]{siunitx} 
\usepackage{etoolbox} 
\robustify\bfseries

\NewDocumentCommand{\anote}{}{\makebox[0pt][l]{$^\textnormal{*}$}}

\usepackage[T1]{fontenc}

\usepackage[utf8]{inputenc}

\usepackage{microtype}

\title{How Gender Debiasing Affects Internal Model Representations, \\ and Why It Matters}

\author{Hadas Orgad\textsuperscript{1} \hspace{2em} Seraphina Goldfarb-Tarrant\textsuperscript{2} \hspace{2em} Yonatan Belinkov\textsuperscript{1}\thanks{~~Supported by the Viterbi Fellowship in the Center for Computer Engineering at the Technion.} \\\\  
  \textsuperscript{1}Technion -- Israel Institute of Technology \hspace{2em} 
  \textsuperscript{2}University of Edinburgh \\
  \texttt{orgad.hadas@cs.technion.ac.il} \hspace{1em} \texttt{s.tarrant@ed.ac.uk} \\ \texttt{belinkov@technion.ac.il}
  }

\begin{document}
\maketitle

\begin{abstract}

Common studies of gender bias in NLP focus either on extrinsic bias measured by model performance on a downstream task or on intrinsic bias found in models' internal representations. However, the relationship between extrinsic and intrinsic bias is relatively unknown. In this work, we illuminate this relationship by measuring both quantities together: we debias a model during downstream fine-tuning, which reduces extrinsic bias, and measure the effect on intrinsic bias, which is operationalized as bias extractability with information-theoretic probing. Through experiments on two tasks and multiple bias metrics, we show that our intrinsic bias metric is a better indicator of debiasing than (a contextual adaptation of) the standard WEAT metric, and can also expose cases of superficial debiasing. Our framework provides a comprehensive perspective on bias in NLP models, which can be applied to deploy NLP systems in a more informed manner. \footnote{Our code and model checkpoints are publicly available at \url{https://github.com/technion-cs-nlp/gender\_internal}}

\end{abstract}

\section{Introduction}

\begin{figure}[!ht]
    \centering
    \includegraphics[width=\columnwidth]{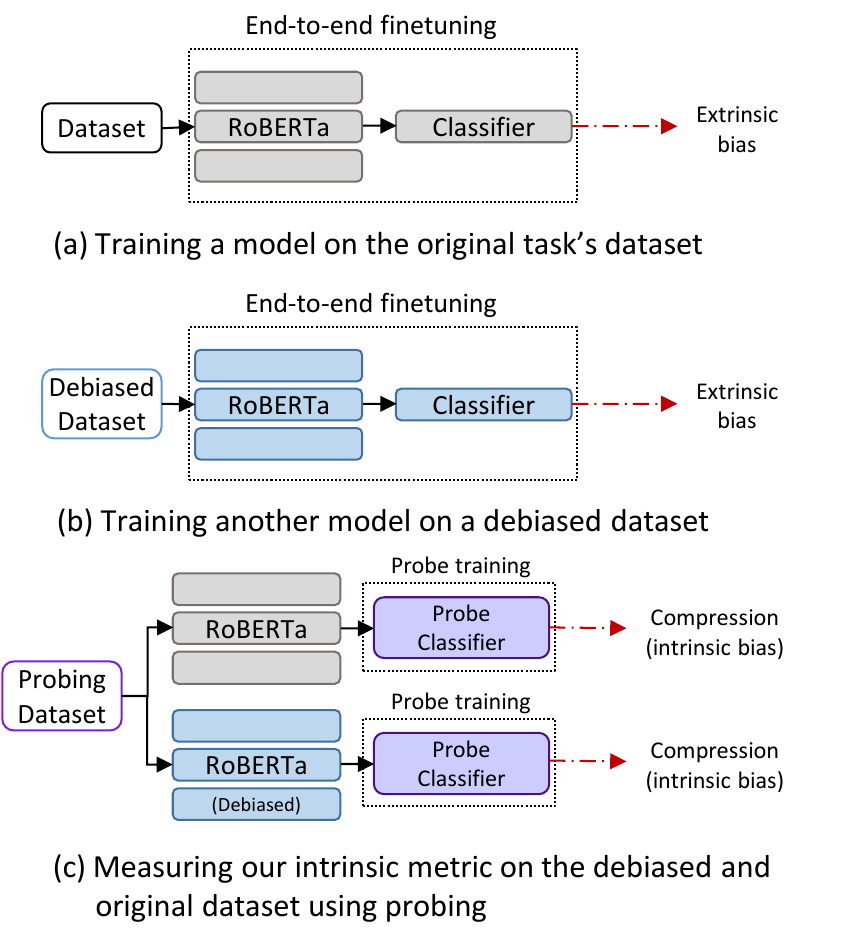}
    \caption{Our proposed framework. Black arrows mark forward passes, red arrows mark things we measure. We first (a) train a model on a downstream task, then (b) train another model on the same task using a debiased dataset, and finally (c) measure intrinsic bias in both models and compare.}
    \label{fig:framework_overview}
\end{figure}

Efforts to identify and mitigate gender bias in Natural Language Processing (NLP) systems typically target one of two notions of bias. \emph{Extrinsic} evaluation methods and debiasing techniques focus on the bias reflected in a downstream task \cite{DeArteaga2019BiasIB, zhao-etal-2018-gender}, while \emph{intrinsic} methods focus on a model's internal representations, such as word or sentence embedding geometry \cite{weat, static-embeddings-bias, guo2021detecting}. Despite an abundance of evidence pointing towards gender bias in pre-trained language models (LMs), the extent of harm caused by these biases is not clear when it is not reflected in a specific downstream task \cite{barocas2017problem, troublewithbias, Blodgett2020LanguageI, bommasani2021opportunities}.
For instance, while the word embedding proximity of ``doctor'' to ``man'' and ``nurse'' to ``woman'' is intuitively normatively wrong, it is not clear when such phenomena would lead to downstream predictions manifesting in social biases. Recently, \citet{intrinsicdonotcorrelate} have shown that debiasing static embeddings intrinsically is not correlated with extrinsic gender bias measures, but the nature of the reverse relationship is unknown: how are extrinsic interventions reflected in intrinsic representations? Furthermore, \citet{lipstick-on-a-pig} demonstrated that a number of intrinsic debiasing methods applied to static embeddings only partially remove the bias and that most of it is still hidden within the embedding. Complementing their view, we examine \emph{extrinsic} debiasing methods, as well as demonstrate the possible harm this could cause. Contrary to their conclusion, we do not claim that these debiasing methods should not be trusted, \emph{as long as they are utilized with care}.

Our goal is to gain a better understanding of the relationship between a model's internal representations and its extrinsic gender bias by examining the effects of various debiasing methods on the model's representations. Specifically, we fine-tune models with and without gender debiasing strategies, evaluate their external bias using various bias metrics, and measure intrinsic bias in the representations. We operationalize intrinsic bias via two metrics: First, we use  CEAT \citep{guo2021detecting}, a contextual adaptation of the widely used intrinsic bias metric WEAT \cite{weat}. Second, we propose to use an information-theoretic probe to quantify the degree to which gender can be extracted from the internal model representations.  Then, we examine how these intrinsic metrics correlate with a variety of extrinsic bias metrics that we measure on the model's downstream performance.  Our approach is visualised in Figure \ref{fig:framework_overview}.

We perform extensive experiments on two downstream tasks (occupation prediction and coreference resolution); several debiasing strategies that involve alterations to the training dataset (such as removing names and gender indicators, or balancing the data by oversampling or downsampling); and a multitude of  extrinsic bias metrics. 
Our analysis reveals new insights into the way language models encode and use information on gender:

\begin{itemize}[itemsep=3pt,parsep=3pt,topsep=3pt]
    \item The effect of debiasing on internal representations is reflected in gender extractability, while not always in CEAT. Thus, gender extractability is a more reliable indicator of gender bias in NLP models.
    \item In cases of high gender extractability but low extrinsic bias metrics, the debiasing is superficial, and the internal representations are a good indicator for this: The bias is still present in internal representations and can be restored by retraining the classification layer. Therefore, our proposed measuring method can help in detecting such cases before deploying the model.
    \item  The two tasks show different patterns of correlation between intrinsic and extrinsic bias. The coreference task exhibits a high correlation. The occupation prediction task exhibits a lower correlation, but it increases after retraining (a case of superficial debiasing). Gender extractability shows higher correlations with extrinsic metrics than CEAT, increasing the confidence in this metric as a reliable measure for gender bias in NLP models.
\end{itemize}

\section{Methodology}

In this study, we investigate the relationship between extrinsic bias metrics of a task and a model's internal representations, under various debiasing conditions, for two datasets in English. We perform extrinsic debiasing, evaluate various extrinsic and intrinsic bias metrics before and after debiasing, and examine correlations.

\paragraph{Dataset.} Let $D = \{\mX, \mY, \mZ\}$ be a dataset consisting of input data $\mX$, labels $\mY$ and protected attributes $\mZ$.\footnote{$\mZ$ is by convention used for attributes for which we want to ensure fairness, such as gender, race, etc. It is purposefully broad, and depending on the task and data could refer to the gender of an entity in coreference, the subject of a text, the demographics of the author of a text, etc.} This work focuses on gender as the protected attribute $z$. In all definitions, $F$ and $M$ indicate female and male gender, respectively, as the value of the protected attribute $z$.

\paragraph{Trained Model.} The model is optimized to solve the downstream task posed by the dataset. It can be formalized as $f \circ g : \mX \rightarrow  \mathbb{R} ^{|\mathcal{Y}|}$, where $g(\cdot)$ is the feature extractor, implemented by a language model, e.g., RoBERTa \cite{roberta}, $f(\cdot)$ is the classification function, and $\mathcal{Y}$ is the set of the possible labels for the task.

\subsection{Bias Metrics}
\label{metrics}
Each bias evaluation method described in the literature can be categorized as extrinsic or intrinsic. In all definitions, $\mathcal{\rvr}$ indicates the model's output probabilities.

\subsubsection{Extrinsic Metrics}

Extrinsic methods involve measuring the bias of a model solving a downstream problem.
The extrinsic metric is a function:
$$E(\mX,\mY,\mR,\mZ) \in \mathbb{R}$$
The output represents the quantity of bias measured; the further from 0 the number is, the larger the bias is. Our analysis comprises a wide range of extrinsic metrics, including some that have been measured in the past on the analyzed tasks \citep{zhao-etal-2018-gender, DeArteaga2019BiasIB, null-it-out, intrinsicdonotcorrelate} and some that have never been measured before, and shows our results apply to many of them. For illustration, we will consider occupation prediction, a common task in research on gender bias \cite{DeArteaga2019BiasIB, null-it-out, romanov2019s}. The input $x$ is a biography and the prediction $y$ is the profession of the person described in it. The protected attribute $z$ is the gender of that person.

\paragraph{Performance gap.} This is the difference in performance metric for two different groups, for instance two groups of binary genders, or a group of pro-stereotypical and a group of anti-stereotypical examples. We measure the following metrics: True Positive Rate (TPR), False Positive Rate (FPR), and Precision. In occupation prediction, for instance, the TPR gap for each profession $y$ expresses the difference in the percentage of women and men whose profession is $y$ and are correctly classified as such. We also measure F1 of three standard clustering metrics for coreference resolution. Each such performance gap captures a different facet of gender bias, and one might be more interested in one of the metrics depending on the application.

We compute two types of performance gap metrics: (1) the sum of absolute gap values over all classes; (2) the Pearson correlation between the performance gap for a class and the percentage of women in that class. For instance, if $y$ is a profession, we measure the correlation between performance gaps and percentages of women in each profession.\footnote{Percentages for coreference resolution are taken from labour statistics, following \citet{zhao-etal-2018-gender}. For occupation prediction we use training set statistics following \citet{DeArteaga2019BiasIB}, \textit{before} balancing.} The two metrics are closely related but  answer slightly different questions: the sum quantifies how a model behaves differently on different genders, and the correlation shows the relation of model behaviour to social biases (in the world or the data) without regard to actual gap size.

\paragraph{Statistical metrics.}
For breadth of analysis, we examine three additional statistical metrics \cite{barocas-hardt-narayanan}, which correspond to different notions of bias.
All three are measured as differences ($d$) between two probability distributions, and we then obtain a single bias quantity per metric by summing all computed distances.
\begin{itemize}[itemsep=2pt,topsep=2pt,parsep=2pt,leftmargin=*]
\item \emph{Independence}:
$d\big(P(\rvr | \rvz = z), P(\rvr)\big) \forall z \in \{F, M\}$.
For instance, we measure the difference between the distribution of model's predictions on women and the distribution of all predictions. Independence is stronger as the prediction $\rvr$ is less correlated with the protected attribute $\rvz$. It is measured with no relation to the gold labels.
\item \emph{Separation}:
$d\big(P(\rvr | \rvy = y, \rvz = z), P(\rvr | \rvy = y)\big)$ $ \forall y \in \mathcal{Y}, z \in \{F, M\}$. For instance, we measure the difference between the distribution of a model's predictions on women who are teachers and the distribution of predictions on all teachers. It encapsulates the TPR and FPR gaps discussed previously, and can be seen as a more general metric.
\item \emph{Sufficiency}:
$d\big(P(\rvy | \rvr = r, \rvz = z), P(\rvy | \rvr = r)\big)$. For instance, we measure the difference between the distribution of gold labels on women classified as teachers by the model and the distribution of gold labels on all individuals classified as teachers by the model. Sufficiency relates to the concept of calibration in classification. A difference in the classifier's scores for men and for women indicates that it might be penalizing or over-promoting one of the genders.
\end{itemize}

\subsubsection{Intrinsic Metrics}

Intrinsic methods are applied to the representation obtained from the feature extractor. These methods are independent of any downstream task.
The intrinsic metric is a function:
$$I(g(\mX), \mZ) \in \mathbb{R} $$

\paragraph{Compression.} Our main intrinsic metric is the \textit{compression} of gender information evaluated by a minimum description length (MDL) probing classifier \cite{voita-titov-2020-information}, trained to predict gender from the model's representations. Probing classifiers are widely used for predicting various properties of interest from frozen model representations  \citep{10.1162/tacl_a_00254}. 
MDL probes were proposed because a probe's accuracy may be misleading due to memorization and other issues \cite{hewitt-liang-2019-designing, probing-belinkov}. We use the MDL online code, where the probe is trained in timesteps, on increasing subsets of the training set, then evaluated against the rest of it. Higher compression indicates greater gender extractability.

\paragraph{CEAT.} We also measure CEAT \citep{guo2021detecting}, which is a contextualized version of WEAT \citep{weat}, a widely used bias metric for static word embeddings. WEAT defines sets $\sX$ and $\sY$ of target words, and sets $\sA$ and $\sB$ of attribute words. For instance, $\sA$ and $\sB$ contain males and females names, while $\sX$ and $\sY$ contain career and family related words, respectively. The bias is operationalized as the geometric proximity between the target and attribute word embeddings, and is quantified in CEAT by the Combined Effect Size (CES) and a p-value for the null hypothesis of having no biased associations. For more information on CEAT refer to Appendix \ref{app:ceat}.

\subsection{Debiasing Techniques}
\label{debiasing-techniques}
We debias models by modifying the downstream task's training data before fine-tuning. 
\emph{Scrubbing} \cite{DeArteaga2019BiasIB} removes first names and gender-specific terms (``he'', ``she'', ``husband'', ``wife'', ``Mr'', ``Mrs'', etc.). \emph{Balancing} subsamples or oversamples examples such that each gender is equally represented in the resulting dataset w.r.t each label. \emph{Anonymization} \cite{zhao-etal-2018-gender} removes named entities. \emph{Counterfactual Augmentation} \cite{zhao-etal-2018-gender} involves replacing male entities in an example with female entities, and adding the modified example to the training set. As some of these are dataset/task-specific, we give more details in the following section. 

\section{Experiments}

\begin{figure}[t]
  \includegraphics[width=\columnwidth]{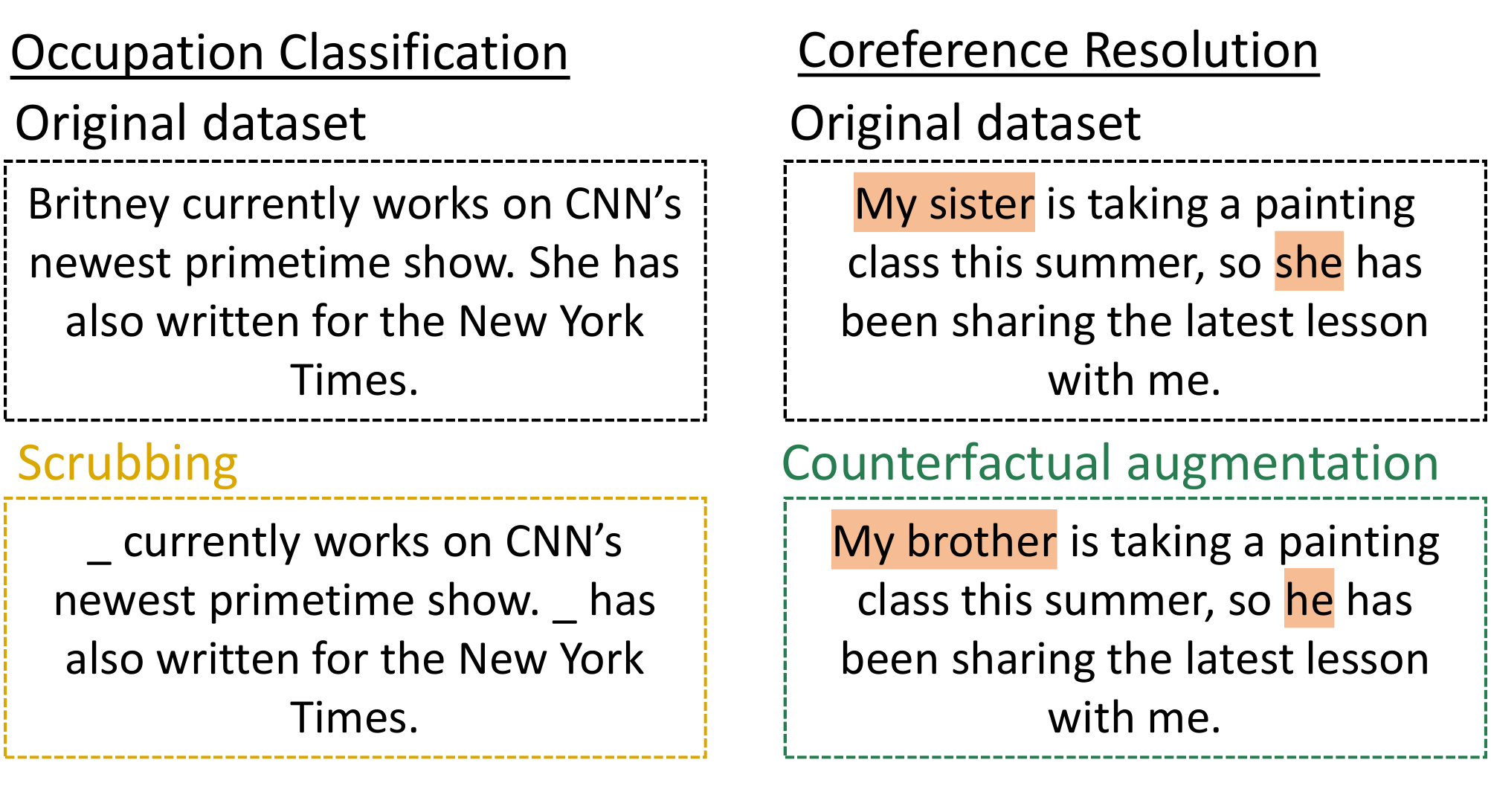}
  \caption{Examples of two debiasing methods performed on the data.}
  \label{fig:tasks}
\end{figure}

In each experiment, we fine-tune a model for a downstream task. For training, we use either the original dataset or a dataset debiased with one of the methods from Section \ref{debiasing-techniques}. Figure \ref{fig:tasks} presents examples of debiasing methods for the two downstream tasks. We measure two intrinsic metrics by probing that model's inner representations for gender extractability (as measured by MDL) and by CEAT, and test various extrinsic metrics. The relation between one intrinsic and one extrinsic metric becomes one data point, and we repeat over many random seeds (for both the model and the probe). Further implementation details are in appendix \ref{app:implementation}.

\subsection{Occupation Prediction}
\label{sec:occupation}
The task of occupation prediction is to predict a person's occupations (from a closed set), based on their biography. We use the Bias in Bios dataset \citep{DeArteaga2019BiasIB}. Regardless of the training method, the test set is subsampled such that each profession has equal gender representation.

\begin{table*}[t]
\centering
\begin{subtable}{\textwidth}
\adjustbox{max width=\textwidth}{%
\sisetup{
            detect-all,
            table-number-alignment = center,
            table-figures-integer = 2,
            table-figures-decimal = 2,
            table-align-text-post = false,
}
\begin{tabular}{lSSSSSSSSSS}
\toprule
&   &   & \multicolumn{8}{c}{\textbf{Extrinsic}} \\
\cmidrule(lr){4-11}
\multirow{2}{*}{\textbf{\shortstack[l]{Debiasing \\ Strategy}}}  & \multicolumn{2}{c}{\textbf{Intrinsic}} & \multicolumn{4}{c}{\textbf{Before}} & \multicolumn{4}{c}{\textbf{After}} \Bstrut \\
\cmidrule(lr){2-3} \cmidrule(lr){4-7} \cmidrule(lr){8-11} \Bstrut
   & {Compression} & {CEAT} &  {TPR (P)}  &  {FPR (S)} &  {Sep} &  {Suff}  & {TPR (P)}   &  {FPR (S)} &  {Sep} &  {Suff} \\
\midrule
Random & 5.61\anote & 0.12$\dagger$ & {-} & {-} & {-} & {-} & {-} & {-} & {-} & {-} \\
Pre-trained & 10.12 & 0.49\anote & {-} & {-} & {-} & {-} & {-} & {-} & {-} & {-} \\
None             & 4.12 & 0.22 & 0.76 & 0.08 & 0.33 & 9.45 & 0.78 & 0.073 & 0.33 & 9.70 \\
Oversampling     & 8.52\anote &  0.29 & 0.73 & 0.09\anote & 0.31 & 8.32\anote & 0.81\anote & 0.068\anote & 0.33 & 10.91\anote\\
Subsampling      & 3.57 & 0.22 & \bfseries 0.32\anote & \bfseries 0.03\anote & \bfseries 0.20\anote & \bfseries 1.22\anote & \bfseries 0.70 \anote & 0.08\anote & 0.30\anote & 1.32\anote\\
Scrubbing        & \bfseries 1.70\anote & 0.23 & 0.70\anote & 0.06\anote & 0.30 & 4.93\anote &  0.71\anote & \bfseries 0.06\anote & \bfseries 2.56\anote & \bfseries 0.81\anote\\
\bottomrule
\end{tabular}}
\caption{Occupation classification: Comparison of intrinsic and extrinsic metrics before and after retraining of classification layer, over 10 seeds per fine-tuned model and per retrained classification model.}
\label{tbl:before-after-bias-in-bios}
\end{subtable}

\medskip

\centering
\begin{subtable}{\textwidth}
\adjustbox{max width=\textwidth}{%
\sisetup{
            detect-all,
            table-number-alignment = center,
            table-figures-integer = 1,
            table-figures-decimal = 2,
            table-align-text-post = false,
}
\begin{tabular}{lSSSSSSSSSS}
\toprule
&   &   & \multicolumn{8}{c}{\textbf{Extrinsic}} \\
\cmidrule(lr){4-11}
\multirow{2}{*}{\textbf{\shortstack[l]{Debiasing \\ Strategy}}} & \multicolumn{2}{c}{\textbf{Intrinsic}}  & \multicolumn{4}{c}{\textbf{Before}} & \multicolumn{4}{c}{\textbf{After}} \Bstrut \\
\cmidrule(lr){2-3} \cmidrule(lr){4-7} \cmidrule(lr){8-11}
  & {Compression} & {CEAT} &  {F1 diff}  & {FPR (S)} &  {Sep} &  {Suff}  & {F1 diff} &  {FPR (S)} &  {Sep} &  {Suff} \\
\midrule
Random      & 0.83\anote            & 0.12$\dagger$      & {-}              & {-}               & {-}                      & {-} & {-} & {-} & {-} & {-} \Tstrut\\
Pre-trained & 0.96            & 0.49\anote   & {-}              & {-}               & {-}                      & {-} & {-} & {-} & {-} & {-} \\
None        & 1.98            & 0.35  & 6.63           & 0.12            & 1.25                   & 8.69 & 6.07 & 0.11 & 1.19 & 7.35 \\
Anon        & 2.07\anote           & 0.31\anote  & 7.26           & 0.13            &  1.34                  & 8.82 &  7.42\anote &  0.13\anote & 1.34\anote & 8.66\anote \\
CA          &  \bfseries 1.50\anote & 0.27\anote   & \bfseries 2.30\anote & 0.05\anote           &  \bfseries 0.54\anote        & 1.67\anote & 3.67\anote & 0.06\anote & 0.67\anote & 2.40\anote \\
Anon + CA   & 1.54\anote           & \bfseries 0.25\anote  & 2.42\anote          & \bfseries 0.049\anote & 0.56\anote & \bfseries 1.56\anote &    \bfseries 2.86\anote & \bfseries 0.05\anote  & \bfseries 0.59\anote &  \bfseries 1.65\anote \\
\bottomrule
\end{tabular}
}
\caption{Coreference resolution: Comparison of intrinsic and extrinsic metrics before and after retraining of classification layer, over 10 seeds per fine-tuned model and 5 seeds per retrained classification model.}
\label{tbl:before-after-coref}
\vspace{-3pt}
\end{subtable}
    \caption{Results on both tasks. * marks significant reduction or increase in bias ($p < 0.05$ on Pitman's permutation test), compared to the non-debiased model (debiasing strategy None). The lowest bias score in each column is marked with \textbf{bold}. P = Pearson; S = Sum. $\dagger$ was computed only on 3 out of 10 tests for which CEAT's $p < 0.05$.}
\label{tbl:before-after-both}
\vspace{-10pt}
\end{table*}

\paragraph{Model.} Our main model is a RoBERTa model \cite{roberta} topped with a linear classifier, which receives the [CLS] token embedding as input and generates a probability distribution over the professions. In addition, we experiment with training a baseline classifier layer on top of a frozen, non-finetuned RoBERTa.  We also replicate our RoBERTa experiments with a DeBERTa model \cite{he2020deberta}, to verify that our results are are not model specific and hold more broadly.

\paragraph{Debiasing Techniques.} Following \citet{DeArteaga2019BiasIB} we experiment with scrubbing the training dataset. Figure~\ref{fig:tasks} shows an example biography snippet and its scrubbed version.  We also conduct balancing (per profession, subsampling and oversampling to ensure an equal number of males and females per profession), which has not previously been used on this dataset and task.

\paragraph{Metrics.} We measure all bias metrics from Section \ref{metrics} except for F1.

\paragraph{Probing.} The probing dataset for this task is the test set, and the gender label of a single biography is the gender of the person described in it. We probe the [CLS] token representation of the biography. In addition to the models described above, we measure baseline extractability of gender information from a randomly initialized RoBERTa model.

\subsection{Coreference Resolution}
\label{sec:coref}
The task of coreference resolution is to find all textual expressions referring to the same real-world entities. We train on Ontonotes 5.0 \citep{Weischedel2017OntoNotesA} and test on the Winobias challenge dataset  \citep{zhao-etal-2018-gender}.  Winobias  consists of sentence pairs, pro- and anti-stereotypical variants, with  individuals referred to by their profession. For example, ``The physician hired the secretary because \textit{he/she} was busy.'' is pro/anti-stereotypical, based on US labor statistics. \footnote{Labor Force Statistics from the Current Population Survey, https://www.bls.gov/cps/cpsaat11.htm} 
A coreference system is measured by the performance gap between the pro- and anti-stereotypical subsets.

\paragraph{Model.} We use the model presented in \citet{lee-etal-2018-higher} with RoBERTa as a feature extractor.

\paragraph{Debiasing Techniques.} Following \citet{zhao-etal-2018-gender}, we apply anonymization (denoted as Anon) and counterfactual augmentation (CA) on the training set. These techniques were used jointly in previous work; we examine each individually as well.

\paragraph{Metrics.} Following \citet{zhao-etal-2018-gender}, we measure the F1 difference between anti- and pro-stereotypical examples.\footnote{We combined the T1 and T2 datasets, as well as the dev and test datasets, to create a single held-out challenge dataset.} We also interpret the task as a classification problem, and measure all metrics from Section \ref{metrics}. For more details refer to Appendix \ref{app:classification_wino}.

\paragraph{Probing.} We probe the representation of a profession word as extracted from Winobias sentences, after masking out the pronouns. We define a profession's gender as the stereotypical gender for this profession. To prevent memorization by the probe---given the small number of professions---the dataset is sorted so that professions are gradually added to the training set, so a success on the validation set is on previously unseen professions.

\section{Results}

Tables \ref{tbl:before-after-bias-in-bios} and \ref{tbl:before-after-coref} present intrinsic and extrinsic metrics for RoBERTa models on the occupation prediction and coreference resolution tasks, respectively. We present a representative subset of the measured metrics that demonstrate the observed phenomena; full results are found in Appendix \ref{app:results}. The DeBERTa model results are consistent with the RoBERTa model trends.

\subsection{Compression Reflects Debiasing Effects}

As shown in the tables, compression  captures differences in models that were debiased differently. CEAT, however, cannot differentiate between occupation prediction models. For example, in occupation prediction (Table~\ref{tbl:before-after-bias-in-bios}) the compression rate varies significantly between a non-debiased and a debiased model via scrubbing and oversampling, while CEAT detects no difference between the models. In coreference resolution (Table~\ref{tbl:before-after-coref}), both compression and CEAT are able to identify differences between the non-debiased model and the others, such as CA, which has both a lower compression and CEAT effect. But the CEAT effect sizes are small (below 0.5), which implies no bias, in contrast to the extrinsic metrics.
\subsection{High Gender Extractability Implies Superficial Debiasing}

\paragraph{Extrinsic and intrinsic effects of debiasing.}
In occupation classification (Table~\ref{tbl:before-after-bias-in-bios}), somewhat surprisingly, subsampling the training data has the strongest effect on extrinsic metrics, but not on compression rate. Scrubbing reduces both intrinsic and extrinsic metrics, although its effect on extrinsic metrics is limited compared to subsampling. Training with oversampling caused less reduction in extrinsic bias metrics. A consequence of oversampling is that some metrics are less biased, but compression rates are increased, so gender information is more accessible. The effectiveness of subsampling over other metrics is further discussed in appendix \ref{app:scrubbing}. In coreference resolution (Table~\ref{tbl:before-after-coref}), while both CA and CA with anonymization reduced gender extractability as well as external bias metrics, anonymization alone \textit{increased} intrinsic bias without affecting external bias metrics significantly.

\begin{table*}[]
\centering
\begin{tabular}{lrrrrrrrr}
\toprule
                & \multicolumn{4}{c}{\textbf{Occupation Classification}} & \multicolumn{4}{c}{\textbf{Coreference Resolution}} \\
                & \multicolumn{2}{c}{$R^2$ Compression} & \multicolumn{2}{c}{$R^2$ CEAT} & \multicolumn{2}{c}{$R^2$ Compression} & \multicolumn{2}{c}{$R^2$ CEAT} \\

\cmidrule(lr){2-5} \cmidrule(lr){6-9}
\textbf{Metric} & Before & After & Before & After & Before & After & Before & After \Bstrut\Tstrut \\
\midrule
F1 diff  ($pro - anti$) & - & - & - & - & \cellcolor[gray]{0.17900000000000005}\color[gray]{1}0.821&\cellcolor[gray]{0.29100000000000004}\color[gray]{1}0.709&\cellcolor[gray]{0.754}\color[gray]{0}0.246&\cellcolor[gray]{0.995}\color[gray]{0}0.005 \\
TPR gap (P) & 
\cellcolor[gray]{0.954}\color[gray]{0}0.046&\cellcolor[gray]{0.696}\color[gray]{0}0.304&\cellcolor[gray]{0.958}\color[gray]{0}0.042&\cellcolor[gray]{0.951}\color[gray]{0}0.049&\cellcolor[gray]{0.778}\color[gray]{0}0.222&\cellcolor[gray]{0.994}\color[gray]{0}0.006&\cellcolor[gray]{0.992}\color[gray]{0}0.008&\cellcolor[gray]{0.988}\color[gray]{0}0.012 \\
TPR gap (S) & 
\cellcolor[gray]{0.951}\color[gray]{0}0.049&\cellcolor[gray]{0.5509999999999999}\color[gray]{0}0.449&\cellcolor[gray]{0.978}\color[gray]{0}0.022&\cellcolor[gray]{0.964}\color[gray]{0}0.036&\cellcolor[gray]{0.18300000000000005}\color[gray]{1}0.817&\cellcolor[gray]{0.248}\color[gray]{1}0.752&\cellcolor[gray]{0.7030000000000001}\color[gray]{0}0.297&\cellcolor[gray]{0.997}\color[gray]{0}0.003 \\
FPR gap (P) & \cellcolor[gray]{0.999}\color[gray]{0}0.001&\cellcolor[gray]{0.88}\color[gray]{0}0.120&\cellcolor[gray]{0.992}\color[gray]{0}0.008&\cellcolor[gray]{0.998}\color[gray]{0}0.002&\cellcolor[gray]{0.979}\color[gray]{0}0.021&\cellcolor[gray]{0.946}\color[gray]{0}0.054&\cellcolor[gray]{0.998}\color[gray]{0}0.002&\cellcolor[gray]{1.0}\color[gray]{0}0.000 \\
FPR gap (S) & \cellcolor[gray]{0.647}\color[gray]{0}0.353&\cellcolor[gray]{0.954}\color[gray]{0}0.046&\cellcolor[gray]{0.921}\color[gray]{0}0.079&\cellcolor[gray]{0.999}\color[gray]{0}0.001&\cellcolor[gray]{0.15600000000000003}\color[gray]{1}0.844&\cellcolor[gray]{0.22699999999999998}\color[gray]{1}0.773&\cellcolor[gray]{0.737}\color[gray]{0}0.263&\cellcolor[gray]{0.996}\color[gray]{0}0.004 \\
Precision gap (P) & 
\cellcolor[gray]{0.996}\color[gray]{0}0.004&\cellcolor[gray]{0.937}\color[gray]{0}0.063&\cellcolor[gray]{0.994}\color[gray]{0}0.006&\cellcolor[gray]{0.998}\color[gray]{0}0.002&\cellcolor[gray]{0.777}\color[gray]{0}0.223&\cellcolor[gray]{0.992}\color[gray]{0}0.008&\cellcolor[gray]{0.991}\color[gray]{0}0.009&\cellcolor[gray]{0.987}\color[gray]{0}0.013 \\
Precision gap (S) & \cellcolor[gray]{0.85}\color[gray]{0}0.150&\cellcolor[gray]{0.7090000000000001}\color[gray]{0}0.291&\cellcolor[gray]{0.969}\color[gray]{0}0.031&\cellcolor[gray]{0.946}\color[gray]{0}0.054&\cellcolor[gray]{0.18300000000000005}\color[gray]{1}0.817&\cellcolor[gray]{0.248}\color[gray]{1}0.752&\cellcolor[gray]{0.704}\color[gray]{0}0.296&\cellcolor[gray]{0.997}\color[gray]{0}0.003 \\
Independence gap (S) &
\cellcolor[gray]{0.749}\color[gray]{0}0.251&\cellcolor[gray]{0.618}\color[gray]{0}0.382&\cellcolor[gray]{0.95}\color[gray]{0}0.050&\cellcolor[gray]{0.995}\color[gray]{0}0.005&\cellcolor[gray]{0.22199999999999998}\color[gray]{1}0.778&\cellcolor[gray]{0.268}\color[gray]{1}0.732&\cellcolor[gray]{0.645}\color[gray]{0}0.355&\cellcolor[gray]{0.999}\color[gray]{0}0.001 \\
Separation gap (S) & 
\cellcolor[gray]{0.9339999999999999}\color[gray]{0}0.066&\cellcolor[gray]{0.835}\color[gray]{0}0.165&\cellcolor[gray]{0.954}\color[gray]{0}0.046&\cellcolor[gray]{0.991}\color[gray]{0}0.009&\cellcolor[gray]{0.16500000000000004}\color[gray]{1}0.835&\cellcolor[gray]{0.22399999999999998}\color[gray]{1}0.776&\cellcolor[gray]{0.739}\color[gray]{0}0.261&\cellcolor[gray]{0.995}\color[gray]{0}0.005 \\
Sufficiency gap (S) & \cellcolor[gray]{0.798}\color[gray]{0}0.202&\cellcolor[gray]{0.433}\color[gray]{1}0.567&\cellcolor[gray]{0.96}\color[gray]{0}0.040&\cellcolor[gray]{0.966}\color[gray]{0}0.034&\cellcolor[gray]{0.17500000000000004}\color[gray]{1}0.825&\cellcolor[gray]{0.247}\color[gray]{1}0.753&\cellcolor[gray]{0.7130000000000001}\color[gray]{0}0.287&\cellcolor[gray]{0.998}\color[gray]{0}0.002 \\
\bottomrule
\end{tabular}
    \caption{Coefficient determination of the regression line taken on the compression rate or CEAT and each extrinsic metric, before and after retraining of the classification layer. P = Pearson; S = Sum.}
    \label{tbl:before-after-r2}
\end{table*}

\paragraph{Debiasing without fine-tuning.}
As the effect on extrinsic bias did not match the effect on intrinsic bias in several cases, we examined the role of the classification layer. We trained a model for occupation prediction without fine-tuning the underlying RoBERTa model. Training on a subsampled dataset also reduced the extrinsic metrics (0.15, 0.03, 0.20, and 0.31, respectively, on TPR gaps Pearson, FPR gaps sum, separation sum, and sufficiency sum). Detailed results of this experiment can be found in Appendix \ref{app:results}. Since no updates were made to the LM, the internal representations could not be debiased, thus the debiasing observed in this model can only be superficial.

\paragraph{Retraining the classification layer.}
Fine-tuning of both tasks revealed that lower extrinsic metrics did not always lead to lower compression. Does this indicate cases where the debiasing process is only superficial, and the internal representations remain biased? To test this hypothesis, we froze the previously fine-tuned LM's weights, and retrained the classification layer. We used the original (non-debiased) training set for retraining.

Tables \ref{tbl:before-after-bias-in-bios} and \ref{tbl:before-after-coref} also compare extrinsic metrics before and after retraining. All models show bias restoration, due to the classification layer being trained on the biased dataset.\footnote{The training datasets contain bias. The occupation prediction set has an unbalanced amount of males and females per profession (for example 15\% of software engineers are females). The coreference resolution training set has more male than female pronouns, and males are more likely to be referred to by their profession \citep{zhao-etal-2018-gender}.} The amount of bias restored varies between models in a way that is predictable by the compression metric.

In the occupation prediction task, comparing Before and After numbers in Table~\ref{tbl:before-after-bias-in-bios}, the model fine-tuned using a scrubbed dataset---which has the lowest compression rate---displays the least bias restoration, confirming that the LM absorbed the process of debiasing. The model fine-tuned on subsampled data has higher extrinsic bias after retraining. Hence, the debiasing was primarily cosmetic, and the representations within the LM were not debiased. The model fine-tuned on oversampled data---which has the highest compression---has the highest extrinsic bias (except for FPR), even though this was not true before retraining.

In coreference resolution, comparing Before and After numbers in Table~\ref{tbl:before-after-coref}, models with the least extrinsic bias (CA and CA+Anon) are also least biased after retraining. Compression rate predicted this; these models also had lower compression rates than non-debiased models. Interestingly, the model fine-tuned with an anonymized dataset is the most biased after retraining, consistent with its high compression rate relative to the other models. As with subsampling and oversampling in occupation prediction, anonymization's (lack of) effect on extrinsic metrics was cosmetic (compare None and Anon in Before block, Table~\ref{tbl:before-after-coref}). Anonymization actually had a biasing effect on the LM, which was realized after retraining.

We conclude that compression rate is a useful indicator of superficial debiasing, and can potentially be used to verify and gain confidence in attempts to debias an NLP model, especially when there is little or no testing data.

\begin{figure*}[t]
\begin{subfigure}[h]{\columnwidth}
  \includegraphics[width=\columnwidth]{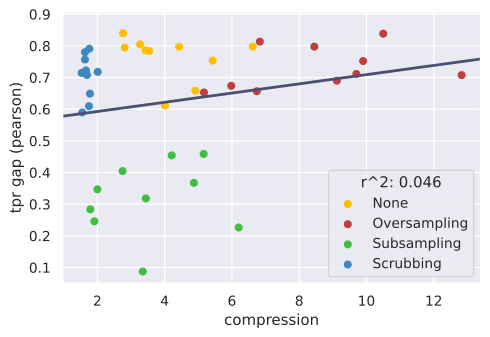}
  \caption{Fine-tuned models. Each point is a single seed for training and testing the model.}
  \label{fig:tpr_bios_corr_compare_before}
\end{subfigure}
\hspace{1em}
\begin{subfigure}[h]{\columnwidth}
  \includegraphics[width=\columnwidth]{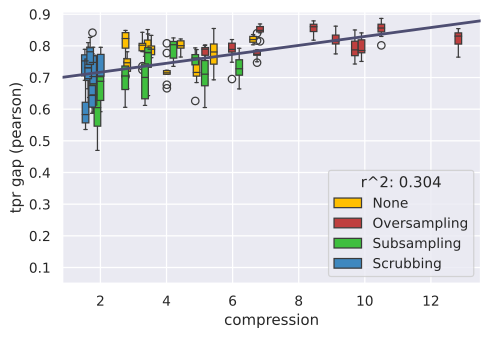}
  \caption{After retraining. Each box represents 10 runs of retraining on the same fine-tuned feature extractor.}
  \label{fig:tpr_bios_corr_compare_after}
\end{subfigure}
\caption{Occupation prediction: Compression vs.\ TPR-gap (Pearson) after various debiasing strategies.}
\label{fig:tpr_bios_corr_compare}
\end{figure*}

\begin{figure*}[t]
\begin{subfigure}[h]{\columnwidth}
  \includegraphics[width=\columnwidth]{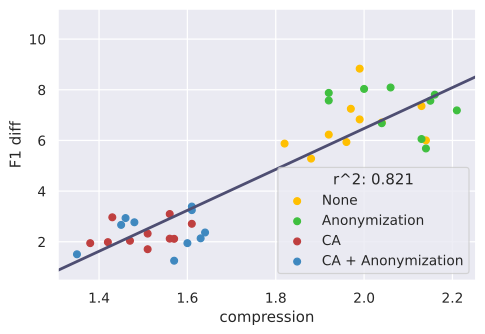}
  \caption{Fine-tuned models. Each point is a single seed for training and testing the model.}
\label{fig:f1_coref_corr_compare_before}
\end{subfigure}
\hspace{1em}
\begin{subfigure}[h]{\columnwidth}
  \includegraphics[width=\columnwidth]{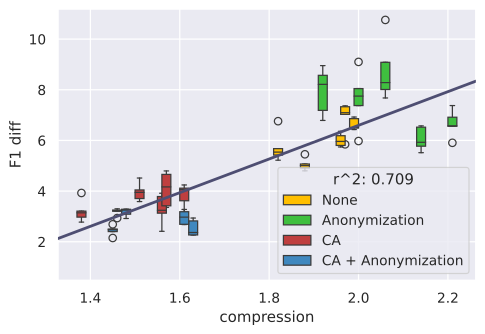}
  \caption{After retraining. Each box represents 5 runs of retraining on the same fine-tuned feature extractor.}
  \label{fig:f1_coref_corr_compare_after}
\end{subfigure}%
\caption{Coreference resolution: Compression vs.\ F1 difference after various debiasing strategies.}
  \label{fig:f1_coref_corr_compare}
\end{figure*}

\subsection{Correlation between Extrinsic and Intrinsic Metrics}

Table \ref{tbl:before-after-r2} shows correlations between compression rate and various extrinsic metrics before and after retraining. In occupation prediction, certain extrinsic metrics have a weak correlation with compression rate, while others do not. Except one metric (FPR gap sum), the compression rate and the extrinsic metric correlate more after retraining. Figure \ref{fig:tpr_bios_corr_compare} illustrates this for TPR-gap (Pearson). The increase is due to superficial debiasing, especially by subsampling data, which prior to retraining had low extrinsic metrics and relatively high intrinsic metrics. This shows that correlation between extrinsic metrics and compression rate for certain metrics is stronger than it appeared before retraining. It is unsurprising that CEAT does not correlate with any extrinsic metrics, since CEAT could not distinguish between different models.

Coreference resolution shows stronger correlations between compression rate and extrinsic metrics, but low correlations between Pearson metrics. We further discuss cases of no correlation in appendix \ref{app:no-corr}. Correlations decrease after retraining, but metrics that were highly correlated remain so ($>0.7$ after retraining). The correlations are visualized for F1 difference metrics in Figure \ref{fig:f1_coref_corr_compare}. CEAT and extrinsic metrics correlate much less than compression rate (Table~\ref{tbl:before-after-r2}). Our results are in line with those of \citet{intrinsicdonotcorrelate}, who found a lack of correlation between extrinsic metrics and WEAT, the static-embedded version of CEAT.

Given that recent work \cite{intrinsicdonotcorrelate,intrinsic_extrinsic_contextual} questions the validity of intrinsic metrics as a reliable indicator for gender bias, the compression rate provides a reliable alternative to current intrinsic metrics, by offering correlation to many extrinsic bias metrics.

\section{Related Work}

There are few studies that examine both intrinsic and extrinsic metrics. Previous work by \citet{intrinsicdonotcorrelate} showed that debiasing static embeddings intrinsically is not correlated with extrinsic bias, challenging the assumption that intrinsic metrics are predictive of bias. We examine the other direction, exploring how extrinsic debiasing affects intrinsic metrics. We also extend beyond their work to contextualized embeddings, a wider range of extrinsic metrics, and a new, more effective intrinsic metric based on information-theoretic probing. A contemporary work by \citet{intrinsic_extrinsic_contextual} measured the correlations between intrinsic and extrinsic metrics in contextualized settings across different language models. In contrast, our work examines the correlations across different versions of the same language model by fine-tuning it using various debiasing techniques.

Studies that inspect extrinsic metrics include either a challenge dataset curated to expose differences in model behavior by gender, or a test dataset labelled by gender.  Among these datasets are Winobias \citep{zhao-etal-2018-gender}, Winogender \citep{rudinger-etal-2018-gender} and GAP \citep{webster-etal-2018-mind} for coreference resolution, WinoMT \citep{stanovsky-etal-2019-evaluating} for machine translation, EEC \citep{kiritchenko-mohammad-2018-examining} for sentiment analysis, BOLD \citep{Dhamala2021BOLDDA} for language generation, gendered NLI \citep{sharma2021evaluating} for natural language inference and Bias in Bios \citep{DeArteaga2019BiasIB} for occupation prediction.

Studies that measure gender bias intrinsically in static word or sentence embeddings measure characteristics of the geometry, such as the proximity between female- and male-related words to stereotypical words, or how embeddings cluster or relate to a gender subspace \citep{static-embeddings-bias, weat,gonen-goldberg-2019-lipstick,ethayarajh-etal-2019-understanding}. However, metrics and debiasing methods  for static embeddings do not apply directly to contextualized ones. Several studies use sentence templates to adapt to contextual embeddings \citep{may-etal-2019-measuring, kurita-etal-2019-measuring, tan-celis-neurips}. This templated approach is difficult to scale, and lacks the range of representations that a contextual embedding offers. Other work extracts embedding representations of words from natural corpora \citep{zhao-etal-2019-gender, guo2021detecting, basta-etal-2019-evaluating}. These studies often adapt the WEAT method \citep{weat}, which measures embedding geometry.
None measure the effect of the presumably found ``bias'' on a downstream task.

There is a growing conversation in the field \cite{barocas2017problem, troublewithbias, Blodgett2020LanguageI, bommasani2021opportunities} about the importance of articulating the harms of measured bias. In general, extrinsic metrics have clear, interpretable impacts for which harm can be defined. Intrinsic metrics have an unclear effect. Without evidence from a concrete downstream task, a found intrinsic bias is only theoretically harmful. Our work is a step towards understanding whether intrinsic metrics provide valuable insights about bias in a model.

\section{Discussion and Conclusions}
\label{discussion}

This study  examined whether bias in internal representations is related to extrinsic bias. We designed a new framework in which we debias a model on a downstream task, and measure its intrinsic bias. We found that gender extractability from internal representations, measured by compression rate via MDL probing, reflects bias in a model. Compression was much more reliable than an alternative intrinsic metric for contextualised representations, CEAT. Compression correlated well---to varying degrees---with many extrinsic metrics. We thus encourage NLP practitioners to use compression as an intrinsic indicator for gender bias in NLP models. When comparing two alternative models, a lower compression rate provides confidence in a model's superiority in terms of gender bias. The relative success of compression over CEAT may be because the compression rate was calculated on the same dataset as the extrinsic metrics, whereas CEAT was measured on a different dataset not necessarily aligned with a specific downstream task. The use of a non-task-aligned dataset is a common strategy among other intrinsic metrics \cite{may-etal-2019-measuring, kurita-etal-2019-measuring, basta2021extensive}. Another possible explanation is that compression rate measures a more focused concept, namely the gender information within the internal representations. CEAT measures proximity among embeddings of general terms that may include other social contexts that do not directly relate to gender (e.g. a female term like `lady' or `Sarah' contains information about not just gender but class, culture, formality, etc, and it can be hard to isolate just one of these from the rest).

Our results show that when a debiasing method reduces extrinsic metrics but not compression, it indicates that the language model remains biased. When such superficial debiasing occurs, the debiased language model may be reapplied to another task, as in \citet{Jin2021Transferability}, resulting in unexpected biases and nullifying the supposed debiasing. Our findings suggest that practitioners of NLP should take special care when adopting previously debiased models and inspect them carefully, perhaps using our framework. Our results differ from those of \citet{mendelson-belinkov-2021-debiasing}, who found that the debiasing increases bias extractability as measured by compression rate. However, they studied different, non-social biases, that arise from spurious or unintended correlations in training datasets (often called dataset biases). In our case, some debiasing strategies increase intrinsic bias while others decrease it. Future work could investigate why debiasing affects extractability differently for these two types of biases.

Our work also highlighted the importance of the classification layer. Using a debiased objective, such as a balanced dataset, the classification layer can provide significant debiasing. This holds even if the internal representations are biased and the classifier is a single linear layer, as shown in the occupation prediction task. Bias stems in part from internal LM bias and in part from classification bias. Practitioners should focus their efforts on both parts when attempting to debias a model.

We used a broader set of extrinsic metrics than is typically used, and found that the bias metrics behaved differently: some decreased more than others after debiasing, and they correlated differently with compression rate. Debiasing efforts may not be fully understood by testing only a few extrinsic metrics. However, compression as an intrinsic bias metric can indicate meaningful debiasing of internal model representations even when not all metrics are easily measurable, since it correlates well with many extrinsic metrics.

A major limitation of this study is the use of gender as a binary variable, which is trans-exclusive. \citet{cao-daume-iii-2020-toward} made the first steps towards inclusive gender bias evaluation in NLP, revealing that coreference systems fail on gender-inclusive text. Further work is required to adjust our framework to non-binary genders, potentially revealing insights about the poor performance of NLP systems in that area.

\section*{Acknowledgements}
This research was supported by the ISRAEL SCIENCE FOUNDATION (grant No. 448/20) and by an Azrieli Foundation Early Career Faculty Fellowship. We also thank Kate McCurdy and Andreas Grivas for comments on early drafts, the members of the Technion CS NLP group for their valuable feedback, and the anonymous reviewers for their useful suggestions.

\bibliographystyle{acl_natbib}
\bibliography{custom, anthology}

\clearpage 

\appendix

\section{Implementation Details}
\label{app:implementation}

We used RoBERTa in all models (base size, 120M parameters). We use  following random seeds in all repeated experiments: 0, 5, 11, 26,  42, 46, 50, 63, 83, 90. Our code was implemented mainly using the Python libraries Pytorch \cite{NEURIPS2019_9015}, Transformers \cite{wolf-etal-2020-transformers}, Sklearn \cite{scikit-learn}, and the experiments were logged using Wandb \cite{wandb}.

\subsection{Occupation Classification}

We fine-tuned a RoBERTa-base model with a linear classification layer on top. Training was done for 10 epochs at a learning rate of 5e-5, batch size of 64. The input to RoBERTa was the biography tokens, which is limited to the first 128 tokens. The resulting [CLS] token embedding is fed to the classifier to predict the occupation. The probing task involves using the same [CLS] token and training the probing classifier to predict the gender of the person in the biography. The experiments without fine-tuning included either a pre-trained or a previously fine-tuned RoBERTa. We first extracted the pre-trained RoBERTa's embeddings of tokens from the [CLS] and then trained a linear classifier on them. The learning rate was 0.001 and the batch size was 64. We trained the classification layer with pre-trained RoBERTa on 300 epochs, but with fine-tuned RoBERTa, 10 epochs were sufficient. For all training processes, the epoch with the greatest validation accuracy was saved. Fine-tuning took 7 hours on a GeForce RTX 2080 Ti GPU. Bias in Bios contains almost 400k biographies, and we obtain validation (10\%) and test set (25\%) by splitting with Scikit-learn's \citep{scikit-learn} test\_train\_split with our random seeds.

\subsection{Coreference Resolution}

We use the implementation of \citet{coref-hoi}, a model that was introduced by \citet{lee2018higher} and has been adopted by many coreference resolution models. Coreference resolution is the process of clustering different mentions in a text that refer to the same real-world entities. The task is solved by detecting mentions through text spans and then predicting for each pair of spans if they represent the same entity. The span representations were extracted with a RoBERTa model, which is fine-tuned throughout the training process, except in the retraining experiment. Fine-tuning took 3 hours on an NVIDIA RTX A6000 GPU. Ontonotes 5.0 has 625k sentences and we use the standard validation and test splits.

\subsection{Probing Classifier}

We use the MDL probe \cite{voita-titov-2020-information} implementation by \citet{mendelson2021debiasing}. In all experiments, we use a linear probe and train it with a batch size of 16 and a learning rate of 1e-3. The timestamps used, meaning the accumulating fractions of data that the probe is trained on, are 2.0\%, 3.0\%, 4.4\%, 6.5\%, 9.5\%, 14.0\%, 21.0\%, 31.0\%, 45.7\%, 67.6\%, 100\%.

\subsection{Metrics}

\subsubsection{Fairness-Based Metrics Implementation}
All three statistical fairness metrics measure the difference between two probability distributions, where this difference describes a notion of bias. We calculate Independence and Separation via Kullback–Leibler (KL) divergence, using the AllenNLP implementation (\url{https://github.com/allenai/allennlp}). We calculate Sufficiency via Wasserstein distance instead, which is motivated by \citet{KwegyirAggrey2021EverythingIR}. In this case, we cannot use KL divergence, since there are some classes that do not occur in model predictions for both male and female genders. This causes the probability distributions to not have the same support, and KL divergence is unbounded. Wasserstein distance lacks the requirement for equal support.

\subsubsection{Classification Metrics Interpretation in Winobias}
\label{app:classification_wino}

Winobias datasets contain pairs of stereotypical and anti-stereotypical sentences. The stereotypes are derived from the US labor statistics (for instance, a profession with a majority of males is stereotypically male). Since coreference resolution is viewed as a clustering problem, it is usually measured via clustering evaluation metrics. Coreference resolution is commonly measured as the average F1 score of these, and the same is true for Winobias. Nevertheless, coreference resolution is accomplished by making a prediction for each pair of mentions, so it can be seen as a classification task. Winobias can be viewed as a simpler task than general coreference resolution, as it contains exactly two mentions of professions and one pronoun, which refers to exactly one profession. Therefore, we reframe it as a classification problem. In a Winobias sentence with two professions $x$ and $y$, as well as a pronoun $p$, where $p$ is referring to $x$, a true positive would be to cluster $x$ and $p$ together, while a false positive would be to cluster $y$ and $p$ together. Our classification metrics are derived based on these definitions. For instance, the TPR gap for profession ``teacher'', which is a stereotypical female occupation, is the TPR rate on pro-stereotypical sentences (with a female pronoun) minus the TPR rate on anti-stereotypical sentences (with a male pronoun).

\subsubsection{CEAT}
\label{app:ceat}

The Word Embedding Association Test (WEAT) developed by \cite{weat} is a method for evaluating bias in static word embeddings. The test is defined as follows: given two sets of target words $\sX, \sY$ (e.g., 'executive', 'management', 'professional' and 'home', 'parents', 'children') and two sets of attribute words (e.g., male names and female names), and using $\Vec{w}$ to represent the word embedding for word $w$, the effect size is:
\begin{equation*}
    \text{ES} = \text{mean}_{x\in \sX} s(x, \sA, \sB) - \text{mean}_{y\in \sY} s(y, \sA, \sB)
\end{equation*}
where
\begin{flalign*}
    &s(x, \sA, \sB) = \\ &\frac{    \text{mean}_{a \in \sA}\text{cos}(\Vec{\vx}, \Vec{\va}) - \text{mean}_{a \in \sA}\text{cos}(\Vec{\vx}, \Vec{\vb})}{\text{std-dev}_{w\in \sX\bigcup \sY}s(w, \sA, \sB)}
\end{flalign*}

In essence, the effect size measures how different are the distances between the embedding vectors of each target group and the attribute groups. Specifically, if $s(x, \sA, \sB) > 0$, $\Vec{\vx}$ is more similar to attribute words $\sB$ and vice versa. For instance, a larger effect size is observed if target words $\sX$ are more similar to attribute words $\sA$ and target words $\sY$ are more similar to attribute words $\sB$.
$|ES| > 0.5$ and $|ES| > 0.8$ are considered medium and large effect sizes, respectively \cite{rice2005comparing}. The null hypothesis holds that there is no difference between the two sets of target words in terms of their relative similarity to the two sets of attribute words, indicating that there are no biased associations. Statistical significance is defined by the p-value of WEAT, which reflects the probability of observing the effect size under the null hypothesis.

 Since a word can take on a great variety of vector representations in a contextual setting, $ES$ varies according to the sentences used to extract word representation. Thus, to adopt WEAT to contextualized representations, the Combined Effect Size (CES) \cite{guo2021detecting} is derived as the distribution of WEAT effect sizes over many possible contextual word representations:
\begin{equation*}
    \text{CES}(\sX,\sY,\sA,\sB) = \frac{\sum_{i=1}^Nv_iES_i}{\sum_{i=1}^Nv_i}
\end{equation*}
where $ES_i$ denotes the WEAT effect size of the $i$'th choice of word representations from a large corpus, and $v_i$ is the inverse of the sum of in-sample variance $V_i$ and between-sample variance in the distribution of random-effects. As in \citet{guo2021detecting}, the representation for each word is derived from 10,000 random sentences extracted from a corpus of Reddit comments.

The combined effect size of each of the models is examined on WEAT stimulus 6, which contains target words of career/family and attribute words of male/female names. This was the only one that detected bias on a pre-trained RoBERTa (CES close to 0.5 and $p<0.05$). The points that we kept in our analysis are those where $p<0.05$, which make up 90\% of the points in occupation prediction and 95\% of the points in coreference resolution.

\section{Full Results}
\label{app:results}
\vspace{-5pt}
In this section we provide the full results of a RoBERTa model trained on the downstream task. The results for the occupation prediction task after fine-tuning are presented in Table \ref{bios-finetuning} and Table \ref{bios-finetuning-retraining} presents the retrained model results. Figure \ref{fig:bios-all} illustrates the correlations between extrinsic metrics and compression rate before and after retraining. Table \ref{bios-no-finetuning} presents the complete results of the model trained without fine-tuning, meaning that the RoBERTa model is the pretrained version from \citet{roberta} and only the classification layer was updated. Subsampling the dataset has significant debiasing effects, which suggests that this debiasing method can achieve low extrinsic bias even when internal bias exists. Table \ref{tbl:deberta} presents the results using a DeBERTa model \cite{he2020deberta}, for the occupation classification task. The trends are similar to those of RoBERTa, with the same metrics showing an increase, no change, or decrease in correlation after re-training, suggesting a general trend in the behavior of these metrics in relation to internal model representations.

Regarding the coreference resolution task, Table \ref{coref-finetuning} displays the results on a finetuned model and Table \ref{coref-finetuning-retraining} displays the retraining results. Figure \ref{fig:coref-all} shows the correlations between compression rate and extrinsic metrics before and after the retraining.

\begin{table*}[h]
\centering
\begin{tabular}{lllll}
\toprule
                & \multicolumn{4}{c}{\textbf{Debiasing Strategy}} \\
\cmidrule{2-5}
\textbf{Metric} & None & Oversampling & Subsampling & Scrubbing \\
\midrule
Compression & 4.121 $\pm$ 1.238 & 8.522* $\pm$ 2.354 & 3.568 $\pm$ 1.516 & \textbf{1.699}* $\pm$ 0.138\\
Accuracy &  \textbf{0.861} $\pm$ 0.005 & 0.852* $\pm$ 0.004 & \textbf{0.861} $\pm$ 0.003 & 0.851* $\pm$ 0.003 \\
TPR gap (P) &  0.763 $\pm$ 0.071 & 0.729 $\pm$ 0.067 & \textbf{0.319}* $\pm$ 0.114 & 0.704* $\pm$ 0.068 \\
TPR gap (S) &  2.391 $\pm$ 0.257 & 2.145* $\pm$ 0.220 & \textbf{1.598}* $\pm$ 0.273 & 2.019* $\pm$ 0.262 \\
FPR gap (P) &  0.591 $\pm$ 0.052 & 0.491* $\pm$ 0.059 & \textbf{0.087}* $\pm$ 0.094 & 0.552 $\pm$ 0.063 \\
FPR gap (S) &  0.075 $\pm$ 0.010 & 0.085* $\pm$ 0.011 & \textbf{0.030}* $\pm$ 0.006 & 0.057* $\pm$ 0.007 \\
Precision gap (P) &  0.398 $\pm$ 0.053 & 0.327* $\pm$ 0.044 & \textbf{0.166}* $\pm$ 0.055 & 0.347* $\pm$ 0.050 \\
Precision gap (S) &   0.015 $\pm$ 0.001 & 0.015 $\pm$ 0.001 & \textbf{0.011}* $\pm$ 0.001 & 0.013* $\pm$ 0.001 \\
Independence gap (S) &  0.009 $\pm$ 0.002 & 0.008 $\pm$ 0.002 & \textbf{0.001}* $\pm$ 0.000 & 0.005* $\pm$ 0.001 \\
Separation gap (S) &  0.327 $\pm$ 0.051 & 0.305 $\pm$ 0.030 & \textbf{0.204}* $\pm$ 0.032 & 0.296 $\pm$ 0.053 \\
Sufficiency gap (S) &  9.451 $\pm$ 1.945 & 8.324* $\pm$ 1.537 & \textbf{1.218}* $\pm$ 0.330 & 4.930* $\pm$ 0.927  \\
\bottomrule
\end{tabular}
    \caption{Occupation Prediction: Results on a RoBERTa-based model trained over 10 seeds. Significant reduction or increase in a metric ($p < 0.05$ on Pitman's permutation test), compared to the non-debiased model (debiasing strategy is None), is marked with *. The lowest bias score or highest performance metric in each column is marked with \textbf{bold}. P = Pearson; S = Sum.}
    
    \label{bios-finetuning}
\end{table*}
\begin{table*}[h]
\centering
\begin{tabular}{lllll}
\toprule
                & \multicolumn{4}{c}{\textbf{Debiasing Strategy}} \\
\cmidrule{2-5}
\textbf{Metric} & None & Oversampling & Subsampling & Scrubbing \\
\midrule
Compression & 4.121 $\pm$ 1.238 & 8.522 $\pm$ 2.354 & 3.568 $\pm$ 1.516 & 1.699 $\pm$ 0.138\\
Accuracy &  0.859 $\pm$ 0.004 & 0.856 $\pm$ 0.003 & 0.853 $\pm$ 0.003 & 0.854 $\pm$ 0.003 \\
TPR gap (P) &  0.777 $\pm$ 0.047 & 0.813* $\pm$ 0.040 & \textbf{0.704}* $\pm$ 0.075 &  0.714* $\pm$ 0.068 \\
TPR gap (S) &   2.482 $\pm$ 0.238 &  2.593* $\pm$ 0.240 & 2.164* $\pm$ 0.284 & \textbf{1.989}* $\pm$ 0.227 \\
FPR gap (P) & 0.596 $\pm$ 0.041 & 0.603 $\pm$ 0.047 & 0.602 $\pm$ 0.041 &  \textbf{0.536}* $\pm$ 0.038 \\
FPR gap (S) &  0.073 $\pm$ 0.008 & 0.068* $\pm$ 0.007 & 0.081* $\pm$ 0.012 &  \textbf{0.059}* $\pm$ 0.005 \\
Precision gap (P) &  0.373 $\pm$ 0.067 &  0.356* $\pm$ 0.070 & 0.338* $\pm$ 0.054 & \textbf{0.309}* $\pm$ 0.053 \\
Precision gap (S) &  0.016 $\pm$ 0.002 & 0.017* $\pm$ 0.002 & 0.015* $\pm$ 0.002 & \textbf{0.014}* $\pm$ 0.002 \\
Independence gap (S) &  0.009 $\pm$ 0.002 &  0.010* $\pm$ 0.002 & 0.009 $\pm$ 0.003 &    \textbf{0.005}* $\pm$ 0.001 \\
Separation gap (S) & 0.334 $\pm$ 0.050 &  0.328 $\pm$ 0.048 &  0.300* $\pm$ 0.049 &  \textbf{0.274}* $\pm$ 0.041 \\
Sufficiency gap (S) &  9.701 $\pm$ 1.305 & 10.908* $\pm$ 1.354 &  8.370* $\pm$ 2.558 & \textbf{5.239}* $\pm$ 0.798 \\
\bottomrule
\end{tabular}
    \caption{Occupation Prediction after retraining: Results on a RoBERTa-based model after retraining of the classification layer with 10 seeds for each pre-trained model. Significant reduction or increase in a metric ($p < 0.05$ on Pitman's permutation test), compared to the non-debiased model (debiasing strategy is None), is marked with *. The lowest bias score or highest performance metric in each column is marked with \textbf{bold}. P = Pearson; S = Sum.}
    \label{bios-finetuning-retraining}
\end{table*}
\begin{table*}[t!]
\centering
\begin{tabular}{lllll}
\toprule
                & \multicolumn{4}{c}{\textbf{Debiasing Strategy}} \\
\cmidrule{2-5}
\textbf{Metric} & None & Oversampling & Subsampling & Scrubbing \\
\midrule
Accuracy & 0.824 $\pm$ 0.003 & 0.815* $\pm$ 0.005 & \textbf{0.831}* $\pm$ 0.001 & 0.807* $\pm$ 0.003 \\
TPR gap (P) & 0.839 $\pm$ 0.011 & 0.443* $\pm$ 0.053 & \textbf{0.158}* $\pm$ 0.156 & 0.814 $\pm$ 0.029 \\
TPR gap (S) & 3.088 $\pm$ 0.192 & \textbf{1.545}* $\pm$ 0.177 & 1.621* $\pm$ 0.088 & 3.154 $\pm$ 0.332 \\
FPR gap (P) & 0.598 $\pm$ 0.016 & 0.369* $\pm$ 0.029 & \textbf{0.067}* $\pm$ 0.050 & 0.550* $\pm$ 0.012 \\
FPR gap (S) & 0.087 $\pm$ 0.004 & 0.041* $\pm$ 0.004 & \textbf{0.027}* $\pm$ 0.003 & 0.112* $\pm$ 0.005 \\
Precision gap (P) & 0.476 $\pm$ 0.027 & 0.163* $\pm$ 0.025	& \textbf{0.134}* $\pm$ 0.065 & 0.479 $\pm$ 0.038 \\
Precision gap (S) &  0.017 $\pm$ 0.001 & 0.012* $\pm$ 0.001 & \textbf{0.010}* $\pm$ 0.001 & 0.016* $\pm$ 0.002\\
Independence gap (S) & 0.014* $\pm$ 0.002 & 0.001* $\pm$ 0.000 & \textbf{0.000}* $\pm$ 0.000 & 0.022* $\pm$ 0.001 \\
Separation gap (S) & 0.336* $\pm$ 0.044 & 0.214* $\pm$ 0.038 & \textbf{0.203}* $\pm$ 0.024 & 0.432* $\pm$ 0.048 \\
Sufficiency gap (S) & 12.019* $\pm$ 1.721 & 2.105* $\pm$ 0.576 & \textbf{1.478}* $\pm$ 0.394 & 13.798* $\pm$ 0.966 \\
\bottomrule
\end{tabular}
    \caption{Occupation Prediction: Results on a RoBERTa-based model trained without fine-tuning, over 5 seeds. The compression rate computed on a pre-trained RoBERTa model is 10.122. Significant reduction or increase in a metric ($p < 0.05$ on Pitman's permutation test), compared to the non-debiased model (debiasing strategy is None), is marked with *. The lowest bias score or highest performance metric in each column is marked with \textbf{bold}. P = Pearson; S = Sum.}
    \label{bios-no-finetuning}
\end{table*}
\begin{table*}[]
\centering
\begin{tabular}{lrrrr}
\toprule
                & \multicolumn{2}{c}{$R^2$ Compression} & \multicolumn{2}{c}{$R^2$ CEAT} \\

\cmidrule(lr){2-3}
\textbf{Metric} & Before & After \Bstrut\Tstrut \\
\midrule
TPR gap (P) & \cellcolor[gray]{0.977}\color[gray]{0}0.023&\cellcolor[gray]{0.88}\color[gray]{0}0.120 & \cellcolor[gray]{0.949}\color[gray]{0}0.051&\cellcolor[gray]{0.994}\color[gray]{0}0.006 \\
TPR gap (S) & \cellcolor[gray]{1.0}\color[gray]{0}0.000&\cellcolor[gray]{0.8}\color[gray]{0}0.200 & \cellcolor[gray]{0.964}\color[gray]{0}0.036 & \cellcolor[gray]{0.902}\color[gray]{0}0.098 \\
FPR gap (P) & \cellcolor[gray]{0.917}\color[gray]{0}0.083&\cellcolor[gray]{0.847}\color[gray]{0}0.153 & \cellcolor[gray]{0.879}\color[gray]{0}0.121 & \cellcolor[gray]{0.851}\color[gray]{0}0.149 \\
FPR gap (S) & \cellcolor[gray]{0.945}\color[gray]{0}0.055&\cellcolor[gray]{0.987}\color[gray]{0}0.013 & \cellcolor[gray]{0.991}\color[gray]{0}0.009 & \cellcolor[gray]{0.979}\color[gray]{0}0.021 \\
Precision gap (P) & \cellcolor[gray]{0.935}\color[gray]{0}0.065&\cellcolor[gray]{0.996}\color[gray]{0}0.004 & \cellcolor[gray]{0.85}\color[gray]{0}0.15 & \cellcolor[gray]{0.975}\color[gray]{0}0.025 \\
Precision gap (S) & \cellcolor[gray]{0.917}\color[gray]{0}0.083&\cellcolor[gray]{0.882}\color[gray]{0}0.118 & \cellcolor[gray]{0.973}\color[gray]{0}0.027& \cellcolor[gray]{0.9319999999999999}\color[gray]{0}0.068 \\
Independence gap (S) & \cellcolor[gray]{0.966}\color[gray]{0}0.034&\cellcolor[gray]{0.916}\color[gray]{0}0.084 & \cellcolor[gray]{1.0}\color[gray]{0}0.0 & \cellcolor[gray]{0.946}\color[gray]{0}0.054 \\
Separation gap (S) & \cellcolor[gray]{1.0}\color[gray]{0}0.000&\cellcolor[gray]{0.883}\color[gray]{0}0.117 & \cellcolor[gray]{0.992}\color[gray]{0}0.008 & \cellcolor[gray]{0.991}\color[gray]{0}0.009 \\
Sufficiency gap (S) & \cellcolor[gray]{0.984}\color[gray]{0}0.016&\cellcolor[gray]{0.75}\color[gray]{0}0.250 & \cellcolor[gray]{0.954}\color[gray]{0}0.046 & \cellcolor[gray]{0.958}\color[gray]{0}0.042 \\
\bottomrule
\end{tabular}
    \caption{Results for a DeBERTa model trained on occupation prediction task. Coefficient determination of the regression line taken on the compression rate or CEAT and each extrinsic metric, before and after retraining of the classification layer. P = Pearson; S = Sum. Coefficients are of lower magnitude for DeBERTa than RoBERTa models, but the same trends apply. They largely increase after retraining (save for FPR gap, and a few of the very low magnitude Pearson metrics). The increase after retraining does not apply to CEAT, and the correlations with CEAT are usually lower.}
    \label{tbl:deberta}
\end{table*}
\FloatBarrier
\begin{figure*}[t]

\begin{subfigure}{\columnwidth}
  \includegraphics[width=\columnwidth]{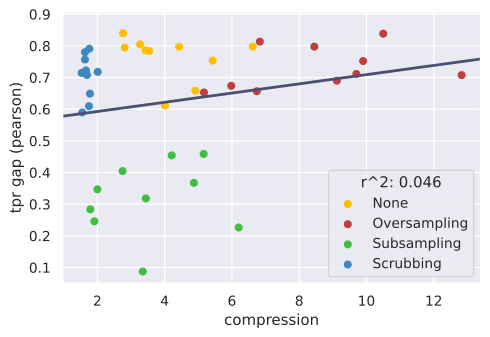}
\end{subfigure}
\hspace{1em}
\begin{subfigure}{\columnwidth}
  \includegraphics[width=\columnwidth]{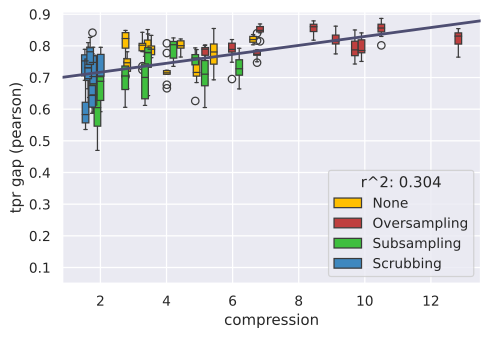}
\end{subfigure}

\begin{subfigure}{\columnwidth}
  \includegraphics[width=\columnwidth]{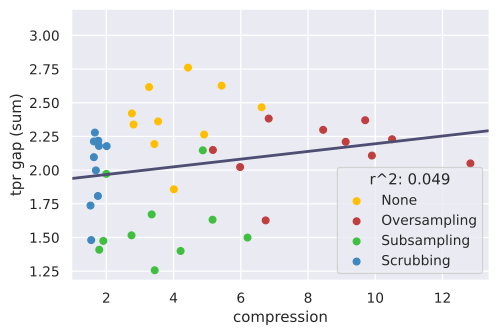}
\end{subfigure}
\hspace{1em}
\begin{subfigure}{\columnwidth}
  \includegraphics[width=\columnwidth]{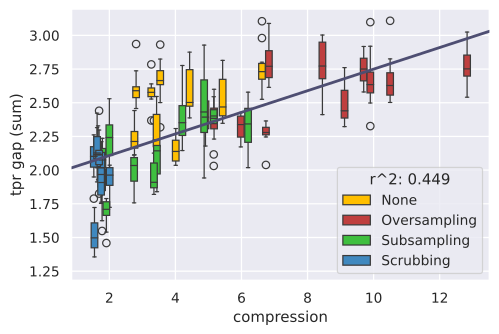}
\end{subfigure}

\begin{subfigure}{\columnwidth}
  \includegraphics[width=\columnwidth]{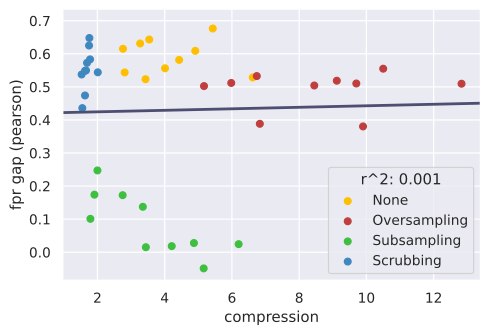}
\end{subfigure}
\hspace{1em}
\begin{subfigure}{\columnwidth}
  \includegraphics[width=\columnwidth]{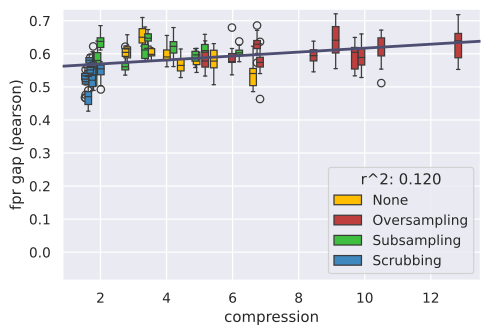}
\end{subfigure}

\begin{subfigure}{\columnwidth}
  \includegraphics[width=\columnwidth]{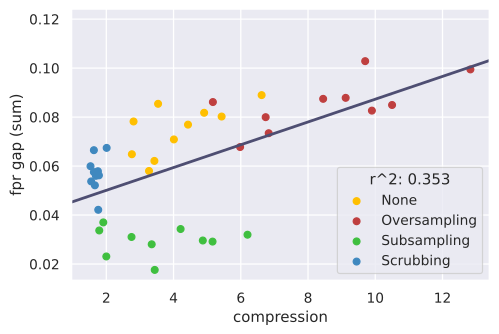}
\end{subfigure}
\hspace{1em}
\begin{subfigure}{\columnwidth}
  \includegraphics[width=\columnwidth]{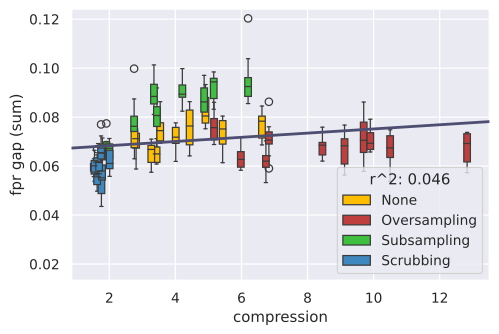}
\end{subfigure}
\caption{Occupation prediction: Before (left) and after (right) plots of compression rate versus and extrinsic metric. Cases of low correlation are discussed in \ref{app:no-corr-occupation}.}

\end{figure*}

\begin{figure*} \ContinuedFloat

\begin{subfigure}{\columnwidth}
  \includegraphics[width=\columnwidth]{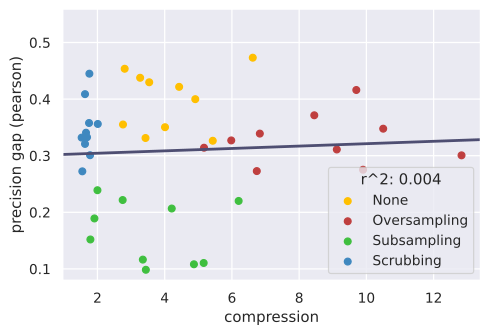}
\end{subfigure}
\hspace{1em}
\begin{subfigure}{\columnwidth}
  \includegraphics[width=\columnwidth]{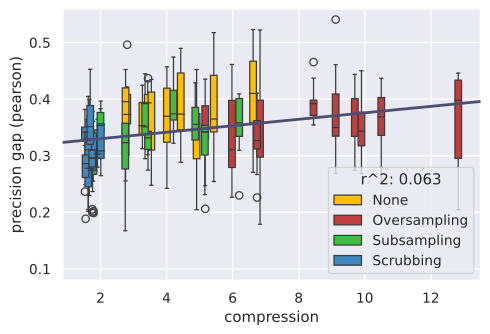}
\end{subfigure}

\end{figure*}

\begin{figure*} \ContinuedFloat

\begin{subfigure}{\columnwidth}
  \includegraphics[width=\columnwidth]{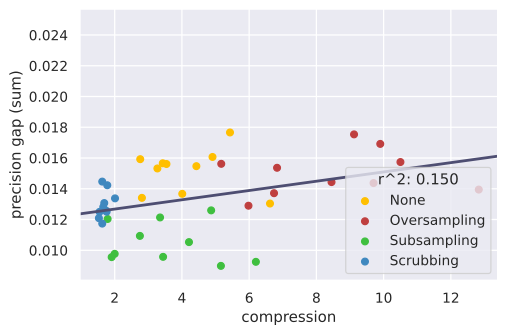}
\end{subfigure}
\hspace{1em}
\begin{subfigure}{\columnwidth}
  \includegraphics[width=\columnwidth]{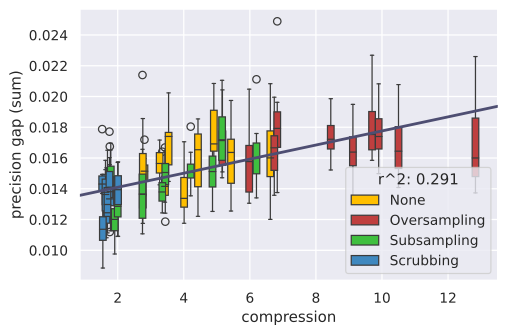}
\end{subfigure}

\begin{subfigure}{\columnwidth}
  \includegraphics[width=\columnwidth]{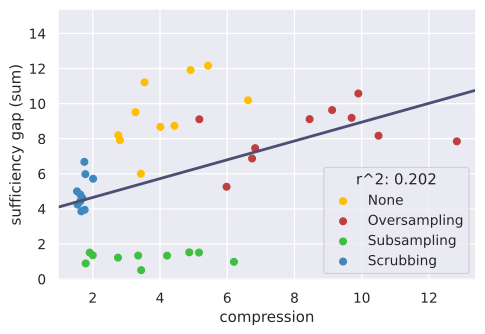}
\end{subfigure}
\hspace{1em}
\begin{subfigure}{\columnwidth}
  \includegraphics[width=\columnwidth]{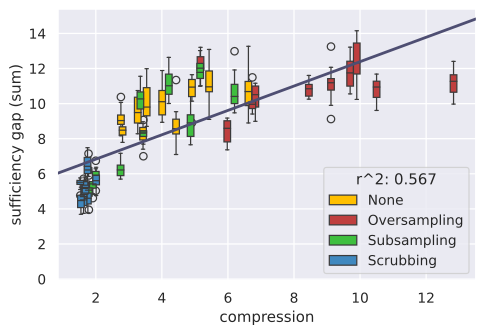}
\end{subfigure}

\begin{subfigure}{\columnwidth}
  \includegraphics[width=\columnwidth]{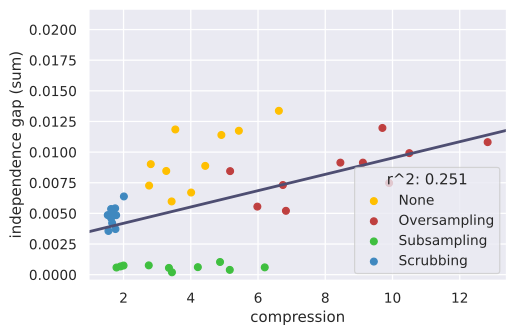}
\end{subfigure}
\hspace{1em}
\begin{subfigure}{\columnwidth}
  \includegraphics[width=\columnwidth]{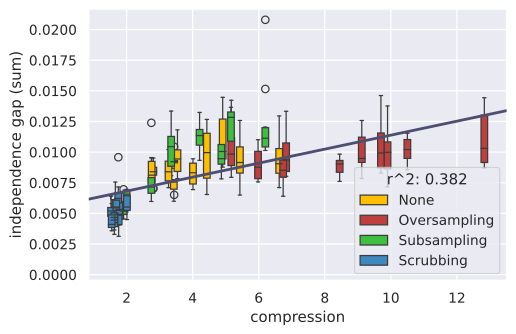}
\end{subfigure}
\caption{Occupation prediction: Before (left) and after (right) plots of compression rate versus and extrinsic metric. Cases of low correlation are discussed in \ref{app:no-corr-occupation}.}

\end{figure*}

\begin{figure*} \ContinuedFloat

\begin{subfigure}{\columnwidth}
  \includegraphics[width=\columnwidth]{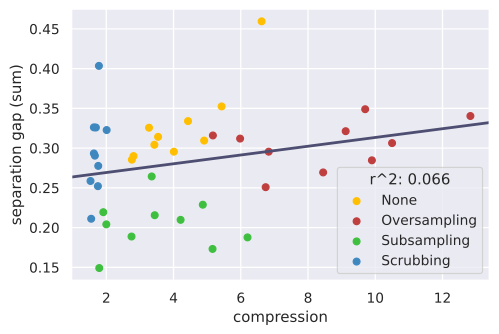}
\end{subfigure}
\hspace{1em}
\begin{subfigure}{\columnwidth}
  \includegraphics[width=\columnwidth]{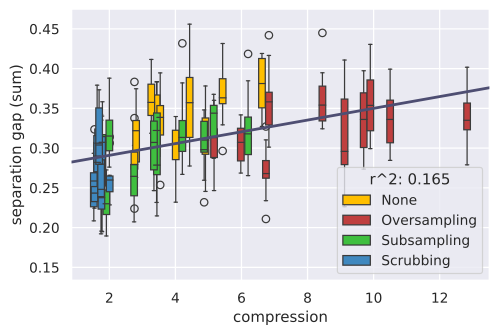}
\end{subfigure}

\caption{Occupation prediction: Before (left) and after (right) plots of compression rate versus and extrinsic metric. Cases of low correlation are discussed in \ref{app:no-corr-occupation}.}
\label{fig:bios-all}

\end{figure*}
\FloatBarrier
\begin{table*}[h]
\centering
\begin{tabular}{lllll}
\toprule
                & \multicolumn{4}{c}{\textbf{Debiasing Strategy}} \\
\cmidrule{2-5}
\textbf{Metric} & None & Anon & CA & Anon + CA \\
\midrule
Compression & 1.984 $\pm$ 0.101 &    2.073* $\pm$ 0.102 & \textbf{1.502}* $\pm$ 0.075 & 1.540* $\pm$ 0.098\\
F1 (Ontonotes test) & 76.406 $\pm$ 0.165 & 76.538 $\pm$ 0.176 & 77.187* $\pm$ 0.071 & \textbf{77.246}* $\pm$ 0.230 \\
F1 diff ($pro - anti$) & 6.631 $\pm$ 1.013 &  7.256 $\pm$ 0.846 & \textbf{2.302}* $\pm$ 0.466 &  2.422* $\pm$ 0.714 \\
TPR gap (P) &  0.654 $\pm$ 0.069 & 0.710* $\pm$ 0.047 & \textbf{0.607} 0.082 $\pm$ & 0.627 $\pm$ 0.100 \\
TPR gap (S) & 4.884 $\pm$ 0.698 & 4.870 $\pm$ 0.509 & 2.041* $\pm$ 0.228 & \textbf{2.014}* $\pm$ 0.286 \\
FPR gap (P) & 0.602 $\pm$ 0.036 & 0.620 $\pm$ 0.056 & \textbf{0.572} $\pm$ 0.078 & 0.629 $\pm$ 0.107 \\
FPR gap (S) &  0.120 $\pm$ 0.015 & 0.128 $\pm$ 0.011 & 0.050* $\pm$ 0.006 & \textbf{0.049}* $\pm$ 0.007 \\
Precision gap (P) & 0.654 $\pm$ 0.068 & 0.710* $\pm$ 0.048& \textbf{0.607} $\pm$	0.083 & 0.627 $\pm$	0.099 \\
Precision gap (S) & 0.061 $\pm$ 0.009 & 0.061 $\pm$ 0.006 & 0.026* $\pm$ 0.003 & \textbf{0.025}* $\pm$ 0.004 \\
Independence gap (S) &  0.027 $\pm$ 0.008 & 0.025 $\pm$ 0.004 & \textbf{0.004}* $\pm$ 0.001 & \textbf{0.004}* $\pm$ 0.001\\
Separation gap (S) & 1.247 $\pm$ 0.150 & 1.344 $\pm$ 0.137 & \textbf{0.537}* $\pm$ 0.061	& 0.557* $\pm$ 0.070\\
Sufficiency gap (S) &  8.684 $\pm$ 1.883 & 8.816 $\pm$ 1.544 & 1.673* $\pm$ 0.354	& \textbf{1.557}* $\pm$ 0.384 \\
\bottomrule
\end{tabular}
    \caption{Coreference resolution: results on Ontonotes test set and Winobias challenge set. Each model was trained over 10 seeds. * Marks significant reduction or increase in bias ($p < 0.05$ on Pitman's permutation test), compared to the non-debiased model (debiasing strategy None). The lowest bias score or highest performance metric in each column is in \textbf{bold}. P = Pearson; S = Sum.}

    \label{coref-finetuning}
\end{table*}

\begin{table*}[h]
\centering
\begin{tabular}{lllll}
\toprule
                & \multicolumn{4}{c}{\textbf{Debiasing Strategy}} \\
\cmidrule{2-5}
\textbf{Metric} & None & Anon & CA & Anon + CA \\
\midrule
Compression & 1.984 $\pm$ 0.065 &    2.073* $\pm$ 0.104& \textbf{1.502}* $\pm$ 0.081 &       1.540* $\pm$ 	0.079\\
F1 (Ontonotes test) & 76.40* $\pm$ 0.16 & 76.48* $\pm$ 0.22 & 76.72* $\pm$ 0.15 & \textbf{76.91}* $\pm$ 0.19 \\
F1 diff ($pro - anti$) &  6.072 $\pm$ 0.789&          7.417* $\pm$ 1.280 & 3.674* $\pm$ 0.599 &               \textbf{2.858}* $\pm$ 0.382 \\
TPR gap (P) &  \textbf{0.635} $\pm$ 0.053 &          0.688* $\pm$ 0.052 & 0.679* $\pm$ 0.062 &               0.654 $\pm$ 0.049\\
TPR gap (S) & 4.561 $\pm$ 0.414 &          5.143* $\pm$ 0.713 & 2.590* $\pm$ 0.420 &               \textbf{2.178}* $\pm$ 0.201 \\
FPR gap (P) & \textbf{0.579} $\pm$ 0.046 &          0.637* $\pm$ 0.055 & 0.620* $\pm$ 0.070 &               0.692* $\pm$ 0.075 \\
FPR gap (S) & 0.113 $\pm$ 0.011 &          0.126* $\pm$ 0.016 & 0.063* $\pm$ 0.010 &               \textbf{0.052}* $\pm$ 0.004 \\
Precision gap (P) &  \textbf{0.636} $\pm$ 0.052 &          0.690* $\pm$ 0.052 & 0.679* $\pm$ 0.062 & 0.652 $\pm$ 0.050 \\
Precision gap (S) & 0.057 $\pm$ 0.005&          0.064* $\pm$ 	0.009 & 0.032* $\pm$ 0.005 &               \textbf{0.027}* $\pm$ 0.003 \\
Independence gap (S) & 0.022 $\pm$ 0.003 &          0.026* $\pm$ 0.006 & 0.006* $\pm$ 0.002 &   0.004* $\pm$ 0.001 \\
Separation gap (S) & 1.188 $\pm$ 0.114 &          1.336* $\pm$ 0.175 & 0.670* $\pm$ 0.111 &      \textbf{0.594}* $\pm$ 0.057 \\
Sufficiency gap (S) &  7.350 $\pm$ 0.914 &          8.655* $\pm$ 1.726 & 0.2.401* $\pm$ 0.610 &               \textbf{1.653}* $\pm$ 0.294 \\
\bottomrule
\end{tabular}
    \caption{Coreference resolution after retraining: results on Ontonotes test set and extrinsic bias metrics on Winobias challenge set. Each model finetuned over 10 seeds and re-trained over 5 seeds. * Marks significant reduction or increase in bias ($p < 0.05$ on Pitman's permutation test), compared to the non-debiased model (debiasing strategy None). The lowest bias score or highest performance metric in each column is in \textbf{bold}. P = Pearson; S = Sum.}
    \label{coref-finetuning-retraining}
\end{table*}

\FloatBarrier
\begin{figure*}[t]
\begin{subfigure}{\columnwidth}
  \includegraphics[width=\columnwidth]{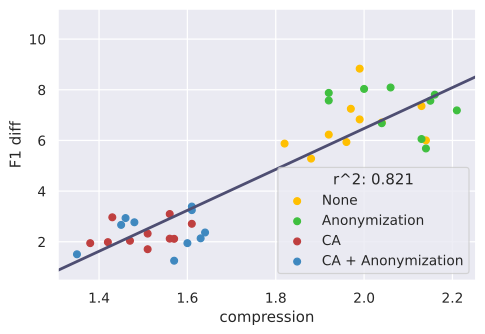}
\end{subfigure}
\hspace{1em}
\begin{subfigure}{\columnwidth}
  \includegraphics[width=\columnwidth]{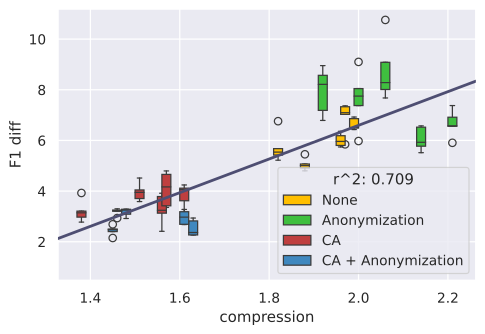}
\end{subfigure}

\begin{subfigure}{\columnwidth}
  \includegraphics[width=\columnwidth]{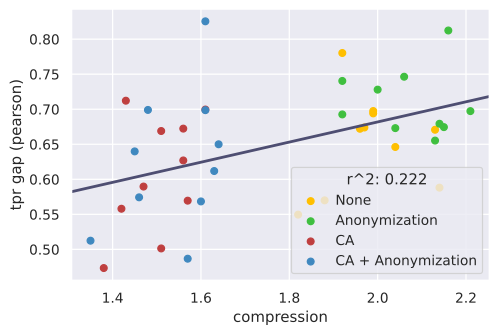}
\end{subfigure}
\hspace{1em}
\begin{subfigure}{\columnwidth}
  \includegraphics[width=\columnwidth]{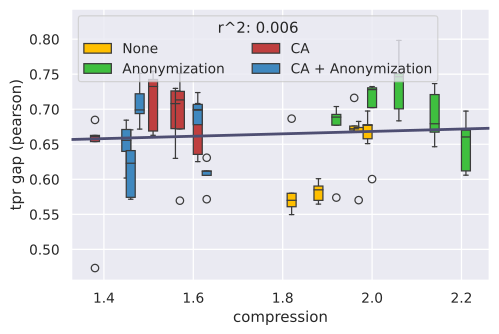}
\end{subfigure}

\begin{subfigure}{\columnwidth}
  \includegraphics[width=\columnwidth]{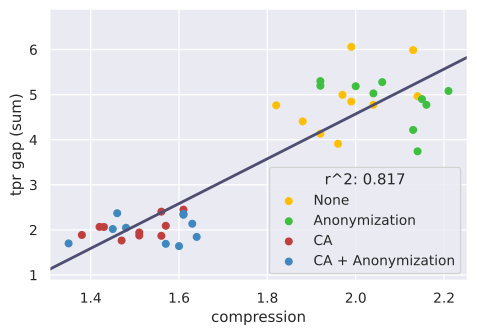}
\end{subfigure}
\hspace{1em}
\begin{subfigure}{\columnwidth}
  \includegraphics[width=\columnwidth]{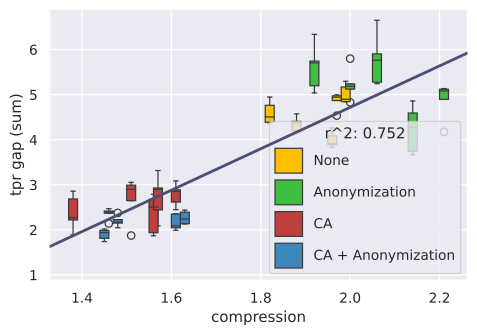}
\end{subfigure}

\begin{subfigure}{\columnwidth}
  \includegraphics[width=\columnwidth]{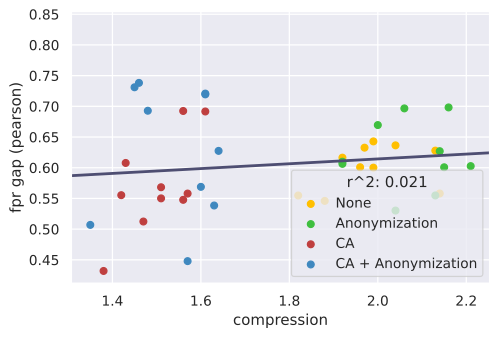}
\end{subfigure}
\hspace{1em}
\begin{subfigure}{\columnwidth}
  \includegraphics[width=\columnwidth]{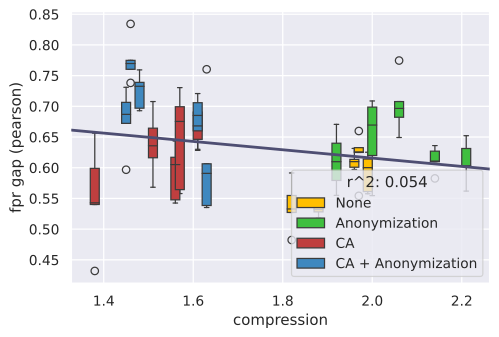}
\end{subfigure}
\caption{Coreference resolution: Before (left) and after (right) plots of compression rate versus and extrinsic metric. Cases of low and no correlation with the Pearson metrics are discussed in \ref{app:no-corr-coreference}.}

\end{figure*}

\begin{figure*} \ContinuedFloat

\begin{subfigure}{\columnwidth}
  \includegraphics[width=\columnwidth]{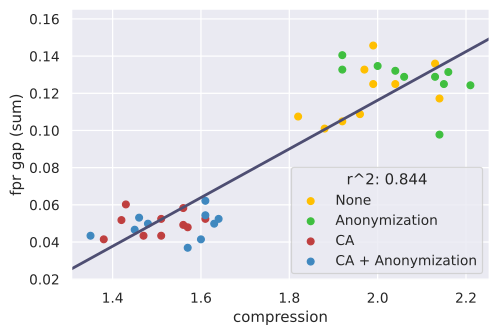}
\end{subfigure}
\hspace{1em}
\begin{subfigure}{\columnwidth}
  \includegraphics[width=\columnwidth]{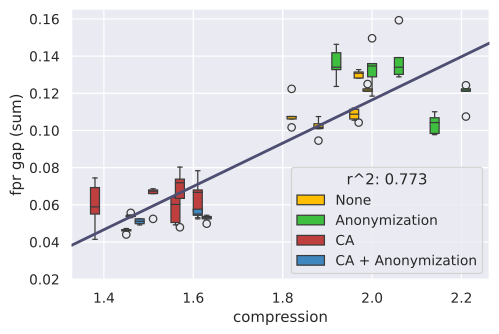}
\end{subfigure}

\begin{subfigure}{\columnwidth}
  \includegraphics[width=\columnwidth]{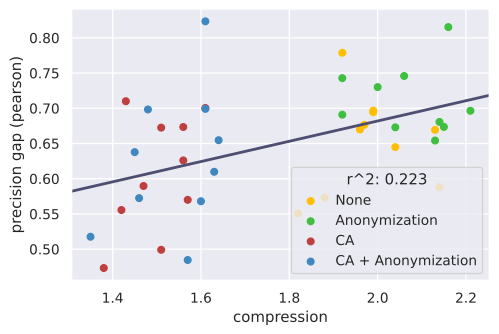}
\end{subfigure}
\hspace{1em}
\begin{subfigure}{\columnwidth}
  \includegraphics[width=\columnwidth]{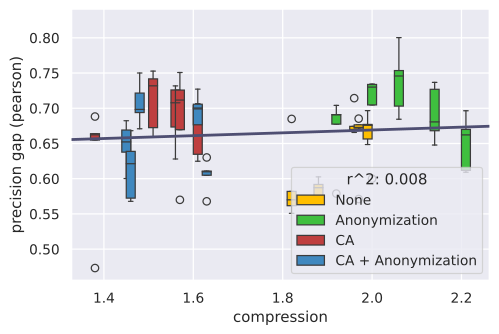}
\end{subfigure}

\begin{subfigure}{\columnwidth}
  \includegraphics[width=\columnwidth]{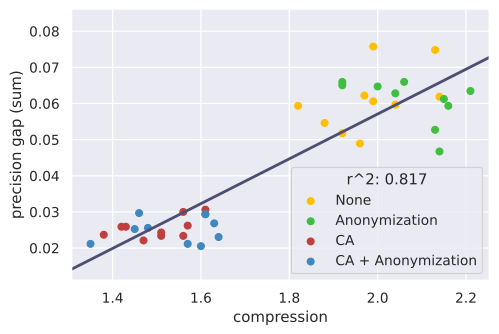}
\end{subfigure}
\hspace{1em}
\begin{subfigure}{\columnwidth}
  \includegraphics[width=\columnwidth]{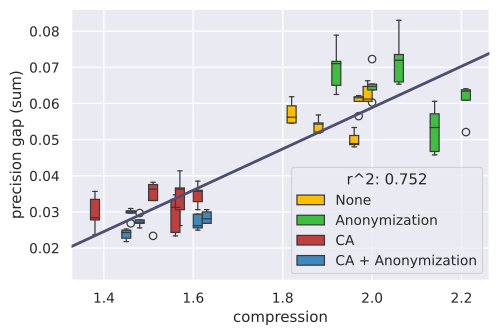}
\end{subfigure}

\begin{subfigure}{\columnwidth}
  \includegraphics[width=\columnwidth]{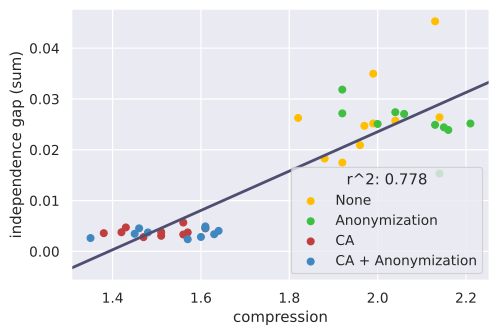}
\end{subfigure}
\hspace{1em}
\begin{subfigure}{\columnwidth}
  \includegraphics[width=\columnwidth]{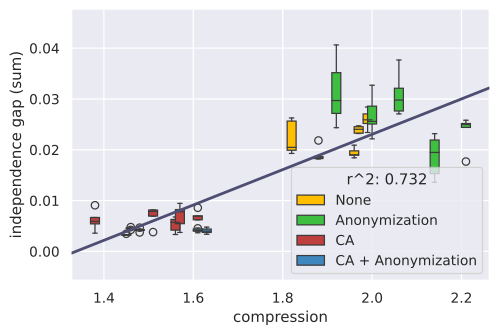}
\end{subfigure}
\caption{Coreference resolution: Before (left) and after (right) plots of compression rate versus and extrinsic metric.}

\end{figure*}

\begin{figure*} \ContinuedFloat

\begin{subfigure}{\columnwidth}
  \includegraphics[width=\columnwidth]{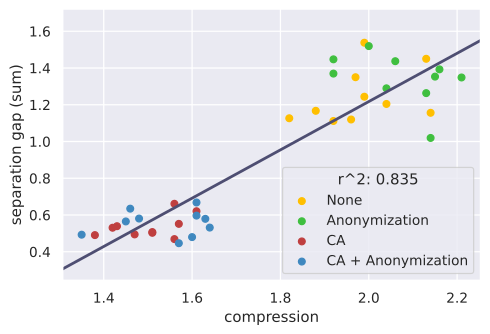}
\end{subfigure}
\hspace{1em}
\begin{subfigure}{\columnwidth}
  \includegraphics[width=\columnwidth]{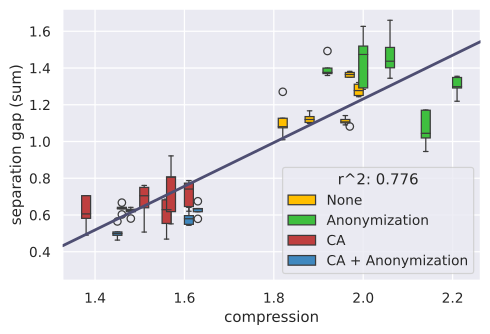}
\end{subfigure}

\begin{subfigure}{\columnwidth}
  \includegraphics[width=\columnwidth]{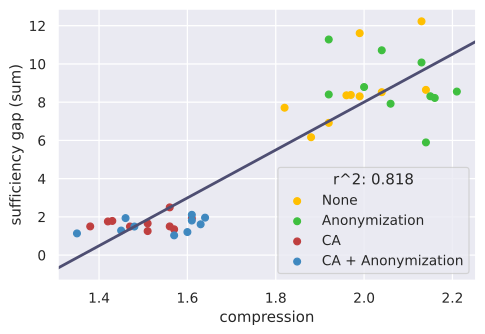}
\end{subfigure}
\hspace{1em}
\begin{subfigure}{\columnwidth}
  \includegraphics[width=\columnwidth]{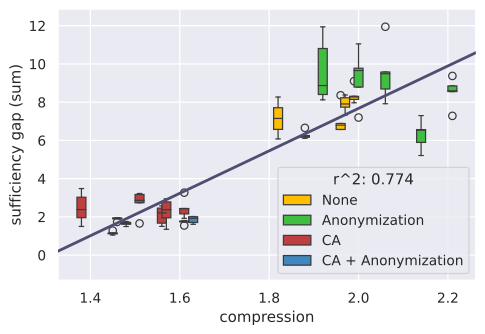}
\end{subfigure}
\caption{Coreference resolution: Before (left) and after (right) plots of compression rate versus and extrinsic metric. Cases of low and no correlation with the Pearson metrics are discussed in \ref{app:no-corr-coreference}.}
\label{fig:coref-all}
\end{figure*}

\FloatBarrier

\begin{table}[h]
\centering
\begin{tabular}{ll}
\toprule
\textbf{Female}   & \textbf{Male} \\
\textbf{Words} & \textbf{Words} \\
\midrule
husband, women, & chief, companies \\
gender, listed, & computer, \\
practices, nurse, & applications, \\
specializes, & md, accepts, \\
children, & known, doctors, \\
ba, child, &  npi, sports,\\
reading, families, &  philosoph',\\
location, place, & problems, rating, \\
affiliated, family, &  no, systems,\\
experiences, &  theory, practicing, \\
spanish, &  software,\\
love, justice &   security, major  \\

\bottomrule
\end{tabular}
    \caption{Top 20 significant words used to predict gender on all biographies, as obtained from a logistic regression model trained on predicting the gender of a person described in a biography. The words are sorted by importance.}
    \label{female-male-words}
\end{table}

\begin{table}[h]
\centering
\begin{tabular}{ll}
\toprule
\textbf{Female}   & \textbf{Male} \\
\textbf{Words} & \textbf{Words} \\
\midrule
husband , women, & holds , emergency, \\
midwife , providing & vanderbilt, forces, \\
book , includes, & registered, mental, \\
joining, faculty &  assistant, president \\

\bottomrule
\end{tabular}
    \caption{Top 8 words used to predict gender of female and male nurses, as obtained from a logistic regression model trained on predicting the gender of a person described in a biography. The words are sorted by importance.}
    \label{nurse-words}
\end{table}

\section{Why is scrubbing not as effective as subsampling?} 
\label{app:scrubbing}

The debiasing method of subsampling significantly reduced external biases in the occupation prediction task. Although compression rates show that scrubbing reduced more gender information, subsampling outperforms it as a debiasing method. We find that in spite of the scrubbing, a probe is able to correctly identify the gender from an internal representation with 68.8\% accuracy compared to 90.7\% on the original, non-scrubbed data. This means that although the scrubbing process reduces extrinsic bias significantly, gender information is still embedded in the [CLS] token embeddings.

To investigate the source of gender information after scrubbing, we use logistic regression (LR) model to predict the gender from the Bag-of-Words of the scrubbed biographies. We perform an iterative process for automatic extra scrubbing: in each iteration we (1) train a LR model for gender prediction (2) scrub the n most significant words for each gender according to the LR weights. The most relevant words among 5 seeds of training with n=10 words scrubbed per iteration are displayed in Table \ref{female-male-words}. The model learns indirect correlations to gender in the absence of explicit gendered words. Because the significant words are related to male- or female-dominated professions, we conducted the process on a specific profession. Table \ref{nurse-words} presents the most significant words for biographies of nurses. There are differences in wording even between females and males in the same profession. The results of this study are in line with the results of other studies that have been conducted on the way biographies are written for men and women \cite{Wagner2016, sun2021men}.

Subsampling is therefore more effective even when gender information is present since it prevents the model from learning correlations between gender information and a profession whereas scrubbing only attempts to remove gender indicators without removing correlations. On the other hand, it is possible that oversampling is less effective for debiasing since seeing more non-unique examples an unrepresented group encourages learning correlations.

\section{A closer look into no-correlation cases}
\label{app:no-corr}
\subsection{Occupation Prediction}
\label{app:no-corr-occupation}

Although compression has the ability to identify bias in most cases, some metrics still show little or no correlation with compression rate. These results suggest that gender information comprises only one facet of embedded bias in the representations. Other factors that may influence these metrics are not considered or measured, such as the connection between a name and a profession.

For example, as can be see in Tables \ref{bios-finetuning} and \ref{bios-finetuning-retraining}, LMs finetuned on subsampled data have the largest FPR gaps after retraining, despite being the least biased before retraining, while those finetuned on oversampled data have the next-to-lowest FPR gaps after retraining. The information encoded in the internal representations may have been encoded in a manner that allowed the classification layer to exhibit a smaller FPR gap when trained on a balanced dataset. However, when the classification layer was retrained on biased training data, it used the same features to make biased predictions.

\subsection{Coreference Resolution}
\label{app:no-corr-coreference}
\begin{figure*}[h]
\centering
\begin{subfigure}{\columnwidth}
  \includegraphics[width=\columnwidth]{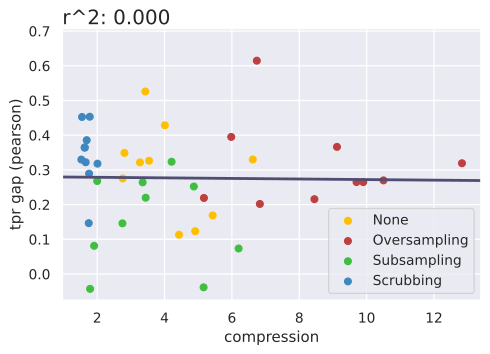}
\end{subfigure}
\hspace{1em}
\begin{subfigure}{\columnwidth}
  \includegraphics[width=\columnwidth]{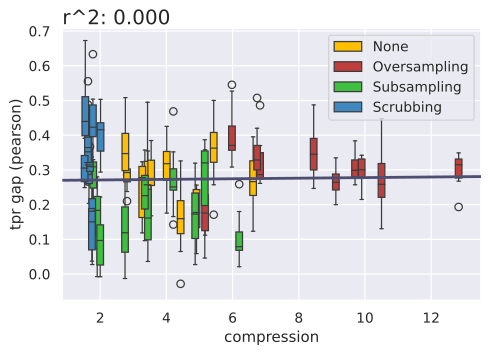}
\end{subfigure}

\begin{subfigure}{\columnwidth}
  \includegraphics[width=\columnwidth]{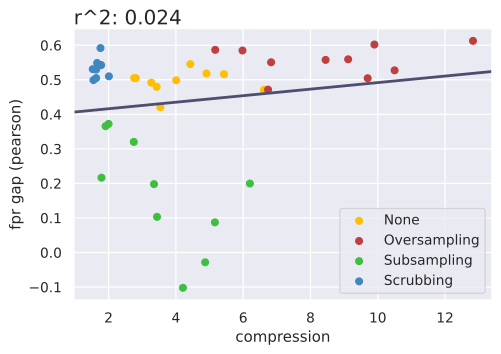}
\end{subfigure}
\hspace{1em}
\begin{subfigure}{\columnwidth}
  \includegraphics[width=\columnwidth]{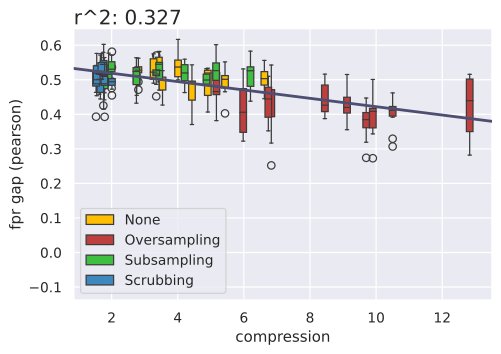}
\end{subfigure}

\begin{subfigure}{\columnwidth}
  \includegraphics[width=\columnwidth]{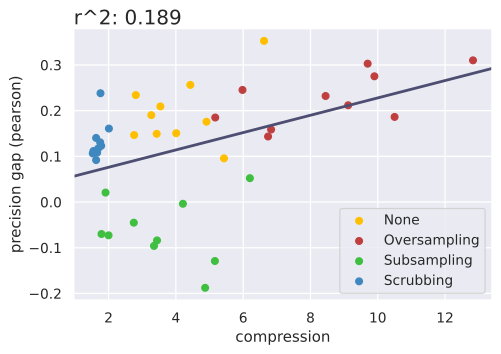}
\end{subfigure}
\hspace{1em}
\begin{subfigure}{\columnwidth}
  \includegraphics[width=\columnwidth]{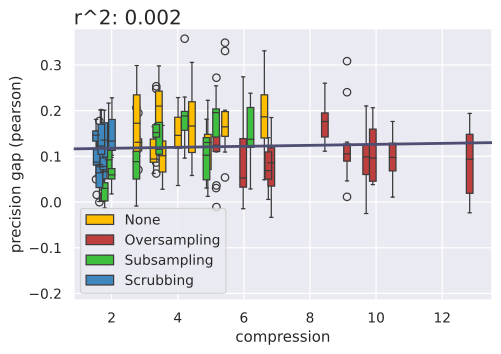}
\end{subfigure}

\caption{Occupation prediction: Before (left) and after (right) plots of compression rate versus Pearson metrics as computed from real-world statistics (as opposed to statistics derived from the training dataset). This shows the unrealiability of using real world statistics to draw conclusions, as they may not be reflected in the data.}
\label{fig:bios-real-world}

\end{figure*}

The cases where there is no correlation between our intrinsic metric and an extrinsic metric are the cases where the metric is based on Pearson correlation. Unlike occupation prediction, coreference resolution seems to exhibit no correlation between those metrics and compression rate. These metrics are computed as the Pearson correlation between a performance gap for a specific profession and the percentage of women in that profession, however the percentages are computed differently in each task: in occupation prediction, the percentages are computed from the train set, focusing on the representation each gender has in the data. In Winobias, the percentages are taken from the US labor statistics, and are unrelated to the training dataset statistics. We note that the two statistics can be different - the real-world representation of women in a profession does not have to be equal to their representation in written text \cite{Suresh2021}. We thus decided to test what happens if we change the statistics used in Winobias to dataset statistics, but Ontonotes 5.0 has very little representation to each profession and the statistics extracted from it would not be reliable. We thus took a different approach and computed the Pearson correlations for occupation prediction with real world statistics instead of dataset statistics. To do this, we mapped the professions appearing in this dataset to professions from the US labor statistics, and dropped those who could no be mapped (6 out of 29 of the professions which is 21.4\%). We then repeated all experiments on the Pearson metrics using these statistics. Figure \ref{fig:bios-real-world} shows the results. Correlations are very different when computed with respect to real-world statistics. TPR-gap has no correlation at all although it had with training data statistics, the correlation for FPR-gap after retraining exists but is negative, and the correlation with precision-gap does not exist after retraining. We thus conclude that the Pearson metrics are less reliable as they are heavily dependent on the statistics with respect to which they are calculated. 

\end{document}